\documentclass[10pt,twocolumn,letterpaper]{article}

\usepackage{cvpr}              

\usepackage{mathrsfs} 
\usepackage{overpic} 
\usepackage{siunitx} 
\usepackage{algorithm} 
\usepackage{algpseudocode} 
\usepackage{etoolbox} 
\usepackage[hang,flushmargin]{footmisc} 
\usepackage[accsupp]{axessibility}  

\newif\ifdraft
\draftfalse

\ifdraft
\newcommand{\itai}[1]{{\color{magenta}[\textbf{Itai:} #1]}}
\newcommand{\harper}[1]{{\color{purple}[\textbf{Harper:} #1]}}
\newcommand{\dale}[1]{{\color{orange}[\textbf{Dale:} #1]}}
\newcommand{\rana}[1]{{\color{cyan}[\textbf{Rana:} #1]}}

\else
\newcommand{\itai}[1]{}
\newcommand{\harper}[1]{}
\newcommand{\dale}[1]{}
\newcommand{\rana}[1]{}

\fi

\newcommand{\ourmethod}{BSB}
\newcommand{\ourtechnique}{Best Segmentation Buddies}

\newcommand{\Ical}{\mathcal{I}}
\newcommand{\Mcal}{\mathcal{M}}

\newcommand{\Vcal}{\mathcal{V}}

\DeclareMathOperator*{\argmax}{\arg\!\max} 
\newcommand{\cossim}{\operatorname{cossim}} 

\definecolor{segblue}{rgb}{0.0,0.21,0.96}
\definecolor{visyellow}{rgb}{0.96,0.69,0.0}
\definecolor{visgreen}{rgb}{0.55, 0.70, 0.50}

\algrenewcommand\algorithmicrequire{\textbf{Input:}} 
\algrenewcommand\algorithmicensure{\textbf{Output:}} 

\definecolor{cvprblue}{rgb}{0.21,0.49,0.74}
\usepackage[pagebackref,breaklinks,colorlinks,allcolors=cvprblue]{hyperref}

\def\paperID{} 
\def\confName{CVPR}
\def\confYear{2026}

\title{\ourtechnique{} for Image-Shape Correspondence}

\author{
\vspace{1mm}
Itai Lang$^{*}$ \hspace{0.9cm} Dongwei Lyu$^{*}$ \hspace{0.9cm} Dale Decatur \hspace{0.9cm} Rana Hanocka \\
University of Chicago \\
{\tt\small \{itailang, dwlyu, ddecatur, ranahanocka\}@uchicago.edu}
}

\begin{document}

\twocolumn[{%
\renewcommand\twocolumn[1][]{#1}%
\maketitle
\begin{center}
    \centering
    \includegraphics[width=0.97\textwidth, trim=10 0 0 10, clip]{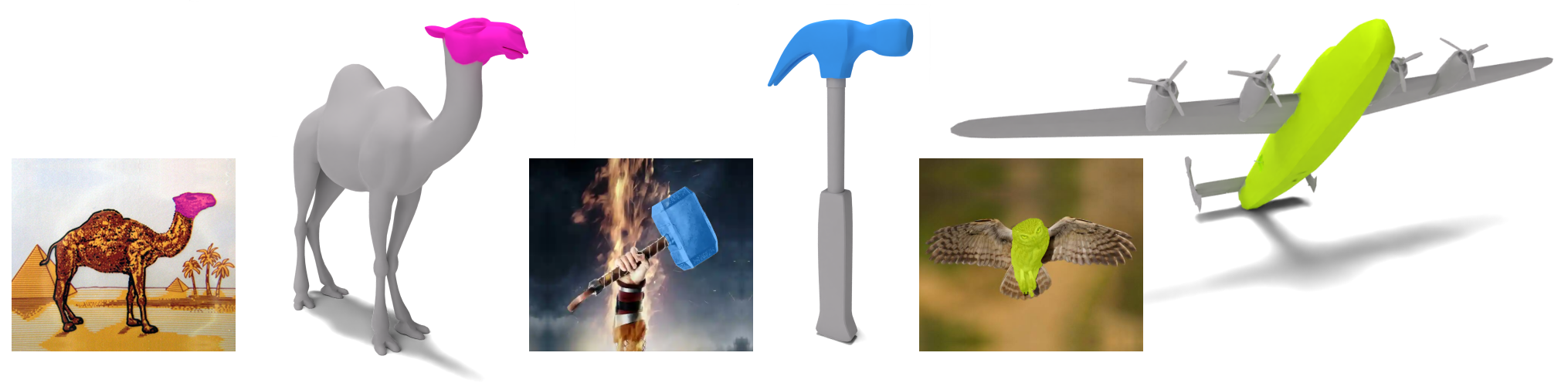}
    \vspace{-4mm}
    \captionof{figure}{Best Segmentation Buddies computes segment-to-segment correspondence across different modalities (image-to-shape) and across different domains. Given an image and a segmentation region, our method finds the corresponding segmentation on a 3D mesh. The image can depict an object similar to the shape (the camel), a part with different geometry (the hammer's head), or even an object from a different domain (the owl and the airplane).}
    \label{fig:teaser}
\end{center}

}]

\def\thefootnote{*}\footnotetext{Equal contribution}


\begin{abstract}
Finding correspondences is a fundamental and extensively researched problem in computer vision and graphics. In this work, we examine the underexplored task of estimating segmentation-to-segmentation correspondence between images in the wild and untextured 3D shapes. This task is highly challenging due to substantial differences in appearance, geometry, and viewpoint. Our approach bridges the cross-modality gap by linking pixels in the image segment to vertices in the corresponding semantic part of the 3D shape. 

To achieve this, we first distill deep visual features from a 2D vision model onto the 3D shape surface, allowing for the computation of feature similarity between image pixels and shape vertices. Then, we identify \textit{Best Segmentation Buddies}, vertices whose most similar image pixel lies within the image segmentation region, enabling the reliable discovery of vertices in semantically corresponding shape parts. Finally, we leverage distilled 3D features from the 2D image segmentation model to segment the shape directly in 3D, bootstrapping the correspondence process. We demonstrate the generality and robustness of our approach across a wide range of image-shape pairs, showcasing accurate and semantically meaningful correspondences. Our project page is at \url{https://threedle.github.io/bsb/}.

\end{abstract}
\vspace{-6pt}  
\section{Introduction} \label{sec:introduction}

Finding correspondences is a long-standing and central problem in computer vision and graphics, with diverse applications such as texture transfer, morphing, and animation. Traditionally, research has focused on within-modality correspondence, \ie, matching shape to shape in the same domain (\eg, human body to human body~\cite{donati2020deepgeometric}).

Recent advances in deep learning have broadened this scope to cross-domain correspondence, but within the same modality. For example, matching an image of a bird to an image of a plane~\cite{aberman2018neural}, or a 3D human to a 3D animal~\cite{abdelreheem2023zero}.
By contrast, correspondence \textit{across modalities} and \textit{across domains}, between 2D images and 3D shapes of different objects, remains largely underexplored. Existing approaches often rely on hard supervision or are tied to specific templates~\cite{neverova2020continuous, shtedritski2024shic}, limiting their versatility and generality.

Establishing correspondence between a \textit{natural image} and an \textit{untextured 3D mesh} is particularly challenging, as these modalities differ drastically in appearance, geometry, and viewpoint. The image contains color and texture, while the shape is textureless and purely geometric. Additionally, the object observed in the image may be substantially different from the 3D shape. Even when depicting the same object, differences in the local geometry (\cref{fig:teaser}, hammer) or pose (\cref{fig:full_shape}, hammer) still exist, making the alignment considerably difficult.

\begin{figure*}
    \centering
    \includegraphics[width=0.99\linewidth, trim=0 10 0 0, clip]{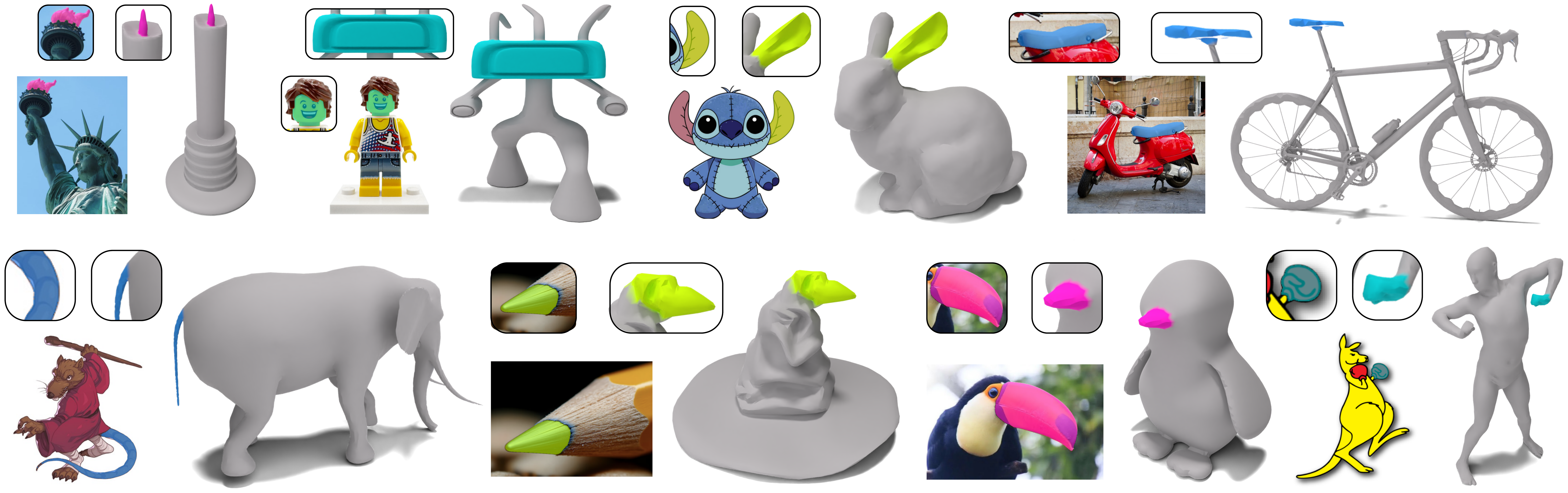}
    \caption{\textbf{Image-shape correspondence gallery.} \ourmethod{} can match semantic parts when the object in the image and the 3D mesh are from \textit{different domains}, where the corresponding elements differ substantially in appearance, shape, and size.}
    \label{fig:gallery}
\end{figure*}

In this work, we address this problem by proposing a \textit{segmentation-to-segmentation} correspondence method across modalities and domains, matching 2D image regions to 3D semantic parts. Unlike sparse keypoint matching \cite{aberman2018neural}, which lacks part semantics, or a restrictive, template-based dense mapping \cite{shtedritski2024shic} between modalities, our approach establishes segmentation-level correspondences. This mid-level representation offers a sweet spot compared to the alternatives, providing more information than sparse landmarks while being robust to local segment variations, revealing meaningful structural relationships between image regions and shape parts.

Our approach builds on the idea of lifting 2D deep visual features \cite{dinov2} onto the 3D surface, enabling direct similarity comparison between image pixels and mesh vertices. However, while deep features can bridge cross-domain gaps~\cite{aberman2018neural}, the modality gap between images and meshes is too large for straightforward matching. To enable cross-modality correspondence, we propose a framework called Best Segmentation Buddies (\ourmethod{}). Instead of enforcing strict mutual nearest neighbors of pixels and vertices in feature space, which rarely align across modalities, we relax the constraint to require matches within the \textit{segmentation region} surrounding the query pixel. This relaxation accommodates feature inconsistencies while maintaining the desired alignment, such that the matched vertex falls within the corresponding semantic 3D part, producing robust correspondences.

To achieve the segmentation-to-segmentation correspondence, we leverage powerful foundation click-based segmentation models~\cite{kirillov2023segment,ravi2024sam2segmentimages} that have proven effective in both 2D and when lifted to 3D \cite{lang2024iseg}. Utilizing the matched vertex, the shape part is obtained using the 3D segmentation model distilled from the 2D one \cite{lang2024iseg}, bootstrapping our cross-modality segmentation-to-segmentation matching between images and shapes.

Notably, our method operates in a zero-shot manner, without any training annotations. It is highly flexible and not restricted to a specific image or shape type, nor to predefined parts. We demonstrate its versatility on a diverse set of images in the wild and different shape categories, spanning organic and manufactured objects, showcasing cross-modality and cross-domain matches. We further show that our method improves over existing zero-shot correspondence baselines, and demonstrate its utility for image-guided local shape texturing.

To summarize, we introduce the problem of segment-to-segment correspondence between natural images and untextured 3D meshes across different domains. To tackle this task, we propose the Best Segmentation Buddies matching scheme, a principled relaxation of mutual nearest neighbors that enforces mutual similarity at the segmentation level to bridge the large cross-modality gap. We show how to combine feature distillation from 2D vision and segmentation foundation models with \ourmethod{} to produce robust correspondences without any training. Through extensive experiments, we show that our approach yields accurate, semantically meaningful matches across diverse image–shape pairs, despite the substantial modality differences.

\section{Related Work} \label{sec:related_work}

\subsection{2D and 3D Segmentation}

A wide range of works focuses on building and using foundational models for 2D segmentation. SAM and SAM2 \cite{kirillov2023segment, ravi2024sam2segmentimages} introduced a promptable architecture capable of zero-shot segmentation. Similarly, SEEM \cite{zou2023segment} provided a unified segmentation interface with visual-language instructions, while SegGPT \cite{seggpt2023} reframed segmentation as a text-to-image generation task and achieved generalization with minimal supervision.

Another line of work explores leveraging the underlying representation (\eg, latent features) learned in pretrained vision models. CLIP and diffusion-based methods \cite{zhou2022extractfreedenselabels, wang2024scliprethinkingselfattentiondense, barsellotti2024diffoff, wang2024diffusionmodelsecretlytrainingfree} generate pseudo-labels based on feature similarity. Self-supervised transformers like DINO and DINOv2 \cite{dino, dinov2} also learn dense visual features that support unsupervised segmentation through clustering or similarity-based grouping. Together, these advances highlight the growing potential of pretrained representations for general-purpose segmentation.

3D segmentation has been extensively studied under both supervised and unsupervised learning paradigms. Supervised methods \cite{qi2017pointnetplusplus, wang2019dynamicgraphcnnlearning, Hanocka_2019} rely on annotated datasets to learn geometric representations, while unsupervised approaches \cite{chen2019baenet, paschalidou2021neuralparts, ren2022extrudenet} leverage autoencoding, geometric priors, or structural regularities to infer part-level structure without labels.

\begin{figure}[!t]
    \centering
    \includegraphics[width=\linewidth, trim=0 0 0 0, clip]{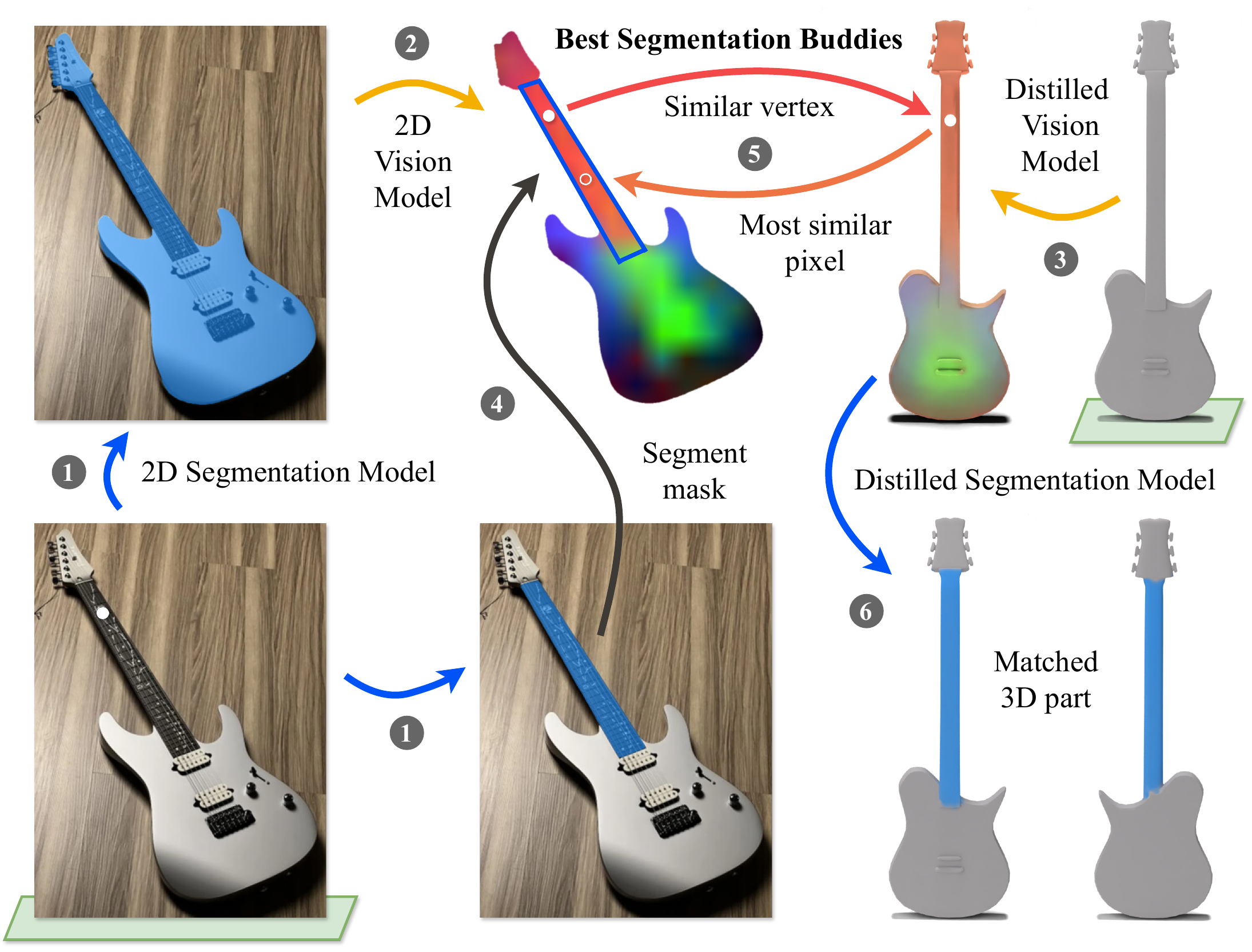}
    \vspace{-6mm}
    \caption{\textbf{Method overview.} Our method finds segmentation-to-segmentation correspondence between an input image (\textcolor{visgreen}{bottom left}) and a 3D mesh (\textcolor{visgreen}{top right}). We connect pixels to mesh vertices by leveraging pretrained 2D foundation models for \textcolor{segblue}{segmentation} (1) and \textcolor{visyellow}{feature similarity} (2). We distill the pretrained vision features to the mesh surface (3) to perform the cross-modality matching (4, 5), and leverage a distilled segmentation model (6) to obtain the corresponding part in 3D.}
    \label{fig:system}
\end{figure}

Recent foundational models such as Point-BERT \cite{yu2022pointbert} pretrain general-purpose 3D representations and enable downstream tasks, including segmentation, but still require task-specific finetuning. Other works \cite{decatur20223dhighlighter, liu2023partslip, abdelreheem2023satr, decatur20223dpaintbrush} leverage pretrained image-text models for zero-shot or few-shot 3D segmentation, but often rely on fixed prompts or offline processing, limiting interactivity and adaptability.

iSeg \cite{lang2024iseg} recently introduced an interactive click-based framework, which allows flexible user-guided segmentation, addressing key limitations in adaptability and usability. In our work, we leverage iSeg to perform click-based segmentation of the 3D shape. Nonetheless, while iSeg and the methods above focus on the segmentation task itself, we use segmentation as a means for computing correspondence.

\subsection{Intra-Modality Correspondence}

Sparse correspondence between images has traditionally relied on keypoint-level matching with hand-crafted descriptors \cite{lowe2004distinctive, calonder2010brief}. Learned descriptors \cite{yi2016liftlearnedinvariantfeature, mishchuk2018workinghardknowneighbors, ebel2019localdescriptors, tian2019secondordersimilarity} such as LIFT improve robustness by optimizing feature mappings through specialized loss functions or network architectures. Transformer-based methods such as COTR \cite{jiang2021cotr} further enable dense, end-to-end matching by leveraging long-range attention.

\begin{figure}[!t]
    \centering
    \includegraphics[width=\linewidth, trim=0 25 0 0, clip]{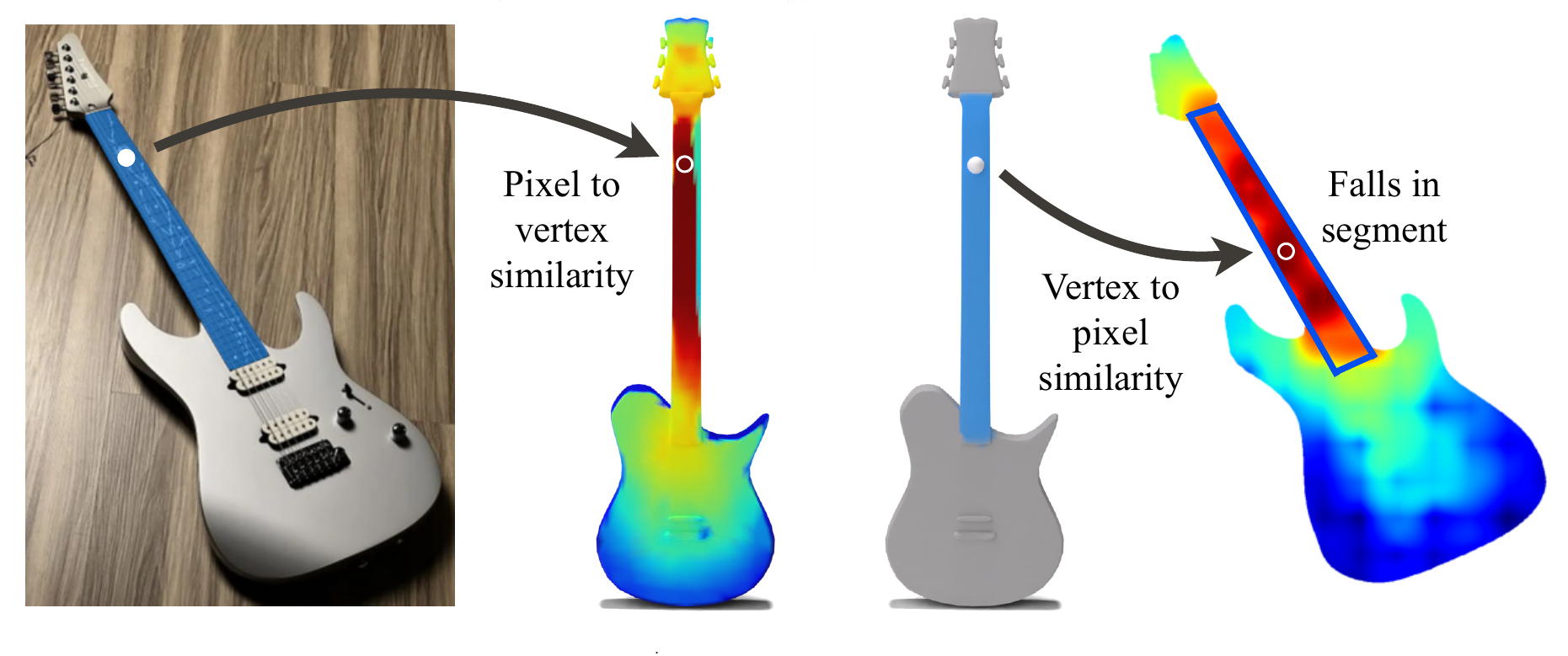}
    \vspace{-5mm}
    \caption{\textbf{\ourtechnique{}.} Pixel to vertex similarity: we visualize the similarity from a clicked pixel feature (left) to the distilled vision features on the mesh (right) with a heatmap (red being most similar and blue being least similar). Vertex to pixel similarity: we visualize the similarity from the distilled feature of the mesh vertex (left) to all the features in the object image region (right). Due to the modality difference, pixels and vertices are not mutual nearest neighbors (a.k.a, best buddies). Thus, for the clicked pixel, we search the best segmentation buddy - a vertex whose nearest neighbor pixel (rightmost) falls within the segmentation region of the original pixel click (leftmost) and yields a segmentation mask with the highest overlap with the click's mask.}
    \label{fig:best_seg_buds}
\end{figure}

More recent approaches, including DIFT \cite{tang2023imagediffusion} and modality inversion in CLIP \cite{mistretta2025crossgap}, demonstrate that off-the-shelf features from pretrained, general-purpose vision models retain sufficient spatial information for correspondence without task-specific supervision. The best buddies strategy \cite{aberman2018neural} complements these methods with a lightweight, symmetric matching mechanism that identifies semantically consistent keypoints directly in the feature space.

Early deep learning methods for shape correspondence adapted CNNs to 3D data via depth maps or volumetric representations \cite{wei2016densehuman, wu20153dshapenet}, but were limited by their Euclidean assumptions and poor deformation handling. Subsequent efforts generalized convolutions to non-Euclidean manifolds using geodesic or spectral constructions \cite{masci2018geodesicconv,boscaini2015,boscaini2016anisotropic}, enabling local descriptor learning on deformable surfaces.

\begin{figure}[!t]
    \centering
    \includegraphics[width=\linewidth, trim=0 40 0 0, clip]{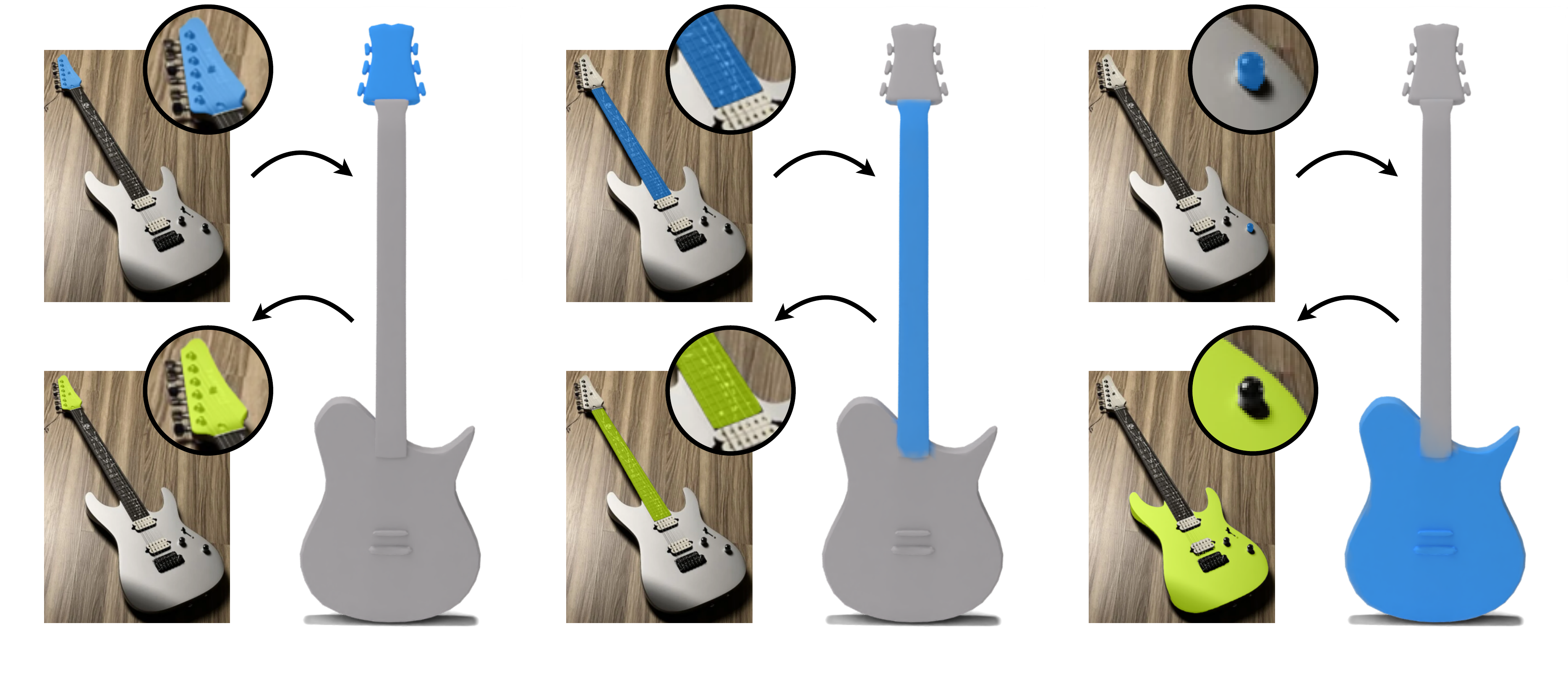}
    \vspace{-5mm}
    \caption{\textbf{\ourtechnique{} matching properties.} When a correspondence between an image region and a mesh part exists (left and middle), the matched vertex will map back to a segment (bottom row) that is almost identical to the original segmentation (top row). However, if a match does not exist, such regions will differ substantially (right), implying the absence of correspondence. We discover this property and exploit it to enable successful image-shape matching via our \ourmethod{} mechanism.}
    \label{fig:bsb_vs_non_bsb}
\end{figure}

Later, the correspondence objective was directly integrated into learning pipelines using functional maps and their unsupervised variants, which enforced geometric consistency \cite{litany2017deep, halimi_2019_CVPR, eisenberger2020deepshell}. However, these methods still assume near-isometric deformations and struggle with diverse topologies. Contrastive learning \cite{donati2020deepgeometric, li2022srfeat} and reconstruction-based frameworks \cite{lang2021dpc,deng2024unsupervised} enabled operating on raw 3D data in an end-to-end pipeline and improved generalization across categories and datasets.

Recent works \cite{abdelreheem2023zero, liu2023openshape} leverage pretrained vision-language models for zero-shot semantic understanding of 3D shapes, though they remain limited to coarse-level alignment due to the granularity of text-supervised representations. In contrast to our work, the above methods all focus on computing correspondences within the same modality, \ie, between pairs of images or 3D shapes, whereas we establish cross-modality matches between 2D images and 3D meshes.

\subsection{Cross-Modality Correspondence}

Prior work in the area of image-to-shape correspondence \cite{Guler_2018_CVPR, Kreiss_2019_CVPR, Rempe_2021_ICCV, pvmaf_2021}, particularly for humans and animals, often targets pose estimation and relies heavily on large annotated datasets. For example, Continuous Surface Embeddings \cite{neverova2020continuous} learns dense correspondences between image pixels and 3D template geometry by training a generalizable pixel and surface encoding utilizing full supervision.

To reduce reliance on annotations, Canonical Surface Mapping methods \cite{discover_rela_2021, kulkarni2019csm, Kulkarni_2020_CVPR} enable inter- and cross-category continuous correspondence between 2D pixels and 3D vertices using self-supervised training with geometric cycle consistency. More recently, SHIC \cite{shtedritski2024shic} eliminates the need for any annotated data by leveraging pretrained vision models, such as DINO and Stable Diffusion, to establish dense point-to-point canonical mappings, yet remains limited in capturing part-level structure and maintaining semantic consistency due to the lack of supervision.

In contrast, our method retains the flexibility of pretrained models while producing segmentation-level correspondences without requiring any annotated data. Furthermore, as far as we are aware, our method is the first approach capable of computing both cross-modality (image-shape) and cross-domain (\eg, owl to airplane) correspondences.

\section{Method} \label{sec:method}

Given a 2D image with a clicked pixel and a 3D textureless mesh, our goal is to find the 3D shape part that semantically matches a segmentation region within the image. We tackle the problem by leveraging click-based 2D and 3D segmentation models \cite{kirillov2023segment, lang2024iseg} that segment an image and a mesh using pixel and vertex clicks, respectively, and ask the question: \textit{how can we match a pixel and a vertex such that their associated segmentations correspond semantically?}

To this end, we propose our novel \ourtechnique{} technique. We assume that a user clicks on a pixel $p$ in the image to get the mask $M^{2D}_p$ using a 2D segmentation model \cite{kirillov2023segment}. Then, we utilize a powerful 2D vision model \cite{oquab2023dinov2} to extract semantic features per pixel, distill the model's information to get semantic features per vertex, and search for a vertex similar to the pixel in the feature space.

However, due to the modality gap between the colored image and the textureless mesh, there is a discrepancy between the image and shape features. Thus, instead of finding a vertex whose most similar pixel is $p$, we find a vertex whose nearest pixel $q^*$ falls within the \textit{segmentation mask} $M^{2D}_p$ associated with the clicked pixel $p$, where the segmented image region $M^{2D}_{q^*}$ for pixel $q^*$ has the highest overlap with $M^{2D}_p$. Such pixel and vertex are considered \textit{\ourtechnique{}}. This measurement aligns the pixel $p$ with a vertex $v_p$ at a semantically matching shape part, which is segmented in turn by the click-based 3D model~\cite{lang2024iseg}. \cref{fig:system,fig:best_seg_buds} depict an overview of our method and the \ourtechnique{} mechanism.

\subsection{Image Segmentation} \label{sec:image_seg}
For a clicked pixel $p$ in an image, we utilize a 2D foundation segmentation model \cite{kirillov2023segment}, which predicts segmentation masks at different granularities. We use two masks: a coarse-scale one $M^{2D}_o$ that delineates an object in the image, and a fine-scale one $M^{2D}_p$ that selects a part of the object. Considering the guitar image in \cref{fig:system} and the click on the guitar's neck, the coarse mask segments the entire guitar, and the fine mask obtains the guitar's neck region.

\subsection{Best Segmentation Buddies} \label{sec:best_seg_buds}
Given the clicked pixel $p$ and its associated part and object masks $M^{2D}_p$ and $M^{2D}_o$, we would like to find mesh vertices whose most similar pixel is inside $M^{2D}_p$. To do so, we compute per-pixel visual features using a 2D vision model $\mathscr{F}^{2D}_{vis}$:
\begin{equation} \label{eq:2D_vis_features}
{F'}_{vis}^{\Ical} = \mathscr{F}^{2D}_{vis}(\Ical) \in \mathbb{R}^{w' \times h' \times d_{vis}},
\end{equation}

\noindent where $\Ical \in [0, 1]^{w \times h \times 3}$ is the input image and $d_{vis}$ the dimension of the visual features. To obtain per-pixel features, we interpolate the output of the vision model to the original image size and get $F_{vis}^{\Ical} \in \mathbb{R}^{w \times h \times d_{vis}}$. These features represent semantic properties of the image pixels.

To find pixel-vertex matches, we should compute semantic features for the mesh vertices in a feature space similar to that of the pixels. Thus, like in iSeg \cite{lang2024iseg}, we lift the 2D vision model features to the 3D surface by rendering the mesh and training a Multi-Layer Perceptron (MLP)  $\mathscr{F}^{3D}_{vis}: \mathbb{R}^3 \rightarrow \mathbb{R}^{d_{vis}}$ that computes deep semantic features per vertex:

\begin{equation} \label{eq:3D_vis_features}
F_{vis}^{\Vcal} = \mathscr{F}^{3D}_{vis}(\Vcal) \in \mathbb{R}^{n \times d_{vis}},
\end{equation}

\noindent where $\Vcal \in \mathbb{R}^{n \times 3}$ are the mesh vertex coordinates, and $n$ is the number of vertices.

Next, we look for vertices that are semantically similar to the clicked image pixel $p$ by computing the cosine similarity between the pixel and the vertex features:
\begin{equation} \label{eq:2D_3D_similarity}
s_{pv} = \frac{{F}_{vis}^{\Ical}[p] \cdot F_{vis}^{\Vcal}[v]}{||{F}_{vis}^{\Ical}[p]||_2 ||F_{vis}^{\Vcal}[v]||_2}.
\end{equation}

Although the shape features are lifted from the 2D vision model used to extract the image features, there is a mismatch between the two feature sets due to the modality gap. First, there is an overall appearance difference between the natural image and the mesh renderings. Second, the 3D features are distilled from multiple views of the shape by averaging the feature information across the different views. Thus, instead of taking the vertex $v$ with the highest cosine similarity to $p$, we consider the $k$ nearest neighbor vertices $\mathcal{C} = \{v'\}$ with the highest similarity to $p$ as candidate matches for the pixel.

Another source of discrepancy between the image and the shape is that the image may contain texture details that are absent from the textureless mesh. Take, for example, the volume knob in the electric guitar image in \cref{fig:bsb_vs_non_bsb}. Such a knob does not appear in the geometry of the guitar shape, and therefore, the knob region in the image cannot be matched to any corresponding part in the 3D mesh.

To cope with this texture mismatch, our \ourtechnique{} come to play. We verify that there are vertices within a semantically matching region in the shape by checking whether their most similar pixel falls inside the segmentation mask of the clicked pixel. Concretely, for each $v' \in \mathcal{C}$, we compute the cosine similarity between the candidate vertex's features and the features of the pixels inside the object mask:
\begin{equation} \label{eq:3D_2D_similarity}
s_{v'q} = \frac{F_{vis}^{\Vcal}[v'] \cdot {F}_{vis}^{\Ical}[q]}{||F_{vis}^{\Vcal}[v']||_2 ||{F}_{vis}^{\Ical}[q]||_2},
\end{equation}

\noindent find the most similar pixel $q' = \argmax_{q \in M^{2D}_o} {s_{v'q}}$, and check if it is inside the mask region of the clicked pixel, namely $q' \in M^{2D}_p$. If this condition holds, the vertex $v'$ is considered to be a best segmentation buddy \textit{candidate} of $p$.

Then, for each best segmentation buddy candidate $v'$, we use the 2D segmentation model with the pixel click $q'$ to obtain the mask $M^{2D}_{q'}$, compute the Intersection over Union (IoU) with the mask of the pixel $p$:
\begin{equation} \label{eq:2D_IoU}
\text{IoU}(M^{2D}_p, M^{2D}_{q'}) = \frac{\text{area}(M^{2D}_p \cap M^{2D}_{q'})}{\text{area}(M^{2D}_p \cup M^{2D}_{q'})}, 
\end{equation}

\noindent and find the pixel with the highest IoU:
\begin{equation} \label{eq:max_IoU_pix}
q^* = \argmax_{q'} {\text{IoU}(M^{2D}_p, M^{2D}_{q'})}.
\end{equation}

\noindent The vertex corresponding to that pixel is denoted $v_p$ and regarded as the \textit{best segmentation buddy} of $p$. Importantly, we note that pixel $p$ is rarely the most similar one to vertex $v_p$. Instead, we require that the nearest-neighbor pixel to $v_p$ has an associated segmentation mask with the maximum IoU with the mask for pixel $p$. Thus, we consider $p$ and $v_p$ to be the most similar to each other in the segmentation sense and coin the term best \textit{segmentation} buddies (\ourmethod{}). A summary of the \ourmethod{} algorithm is provided in the supplementary.

The \ourmethod{} requirement strikes the right balance of restrictiveness. On the one hand, it does not require pixel $p$ to be the nearest neighbor of vertex $v_p$. Instead, the restriction is relaxed to a pixel within the segmentation region associated with $p$, which enables finding vertices in a semantically similar region of the 3D shape while accommodating the image and mesh feature discrepancy due to the modality gap. On the other hand, this mechanism is restrictive enough to filter out texture-related elements in the image that lack a geometric counterpart in the mesh, thereby addressing the texture mismatch between the image and the shape.

\begin{figure}
    \centering
    \includegraphics[width=\linewidth, trim=0 0 0 0, clip]{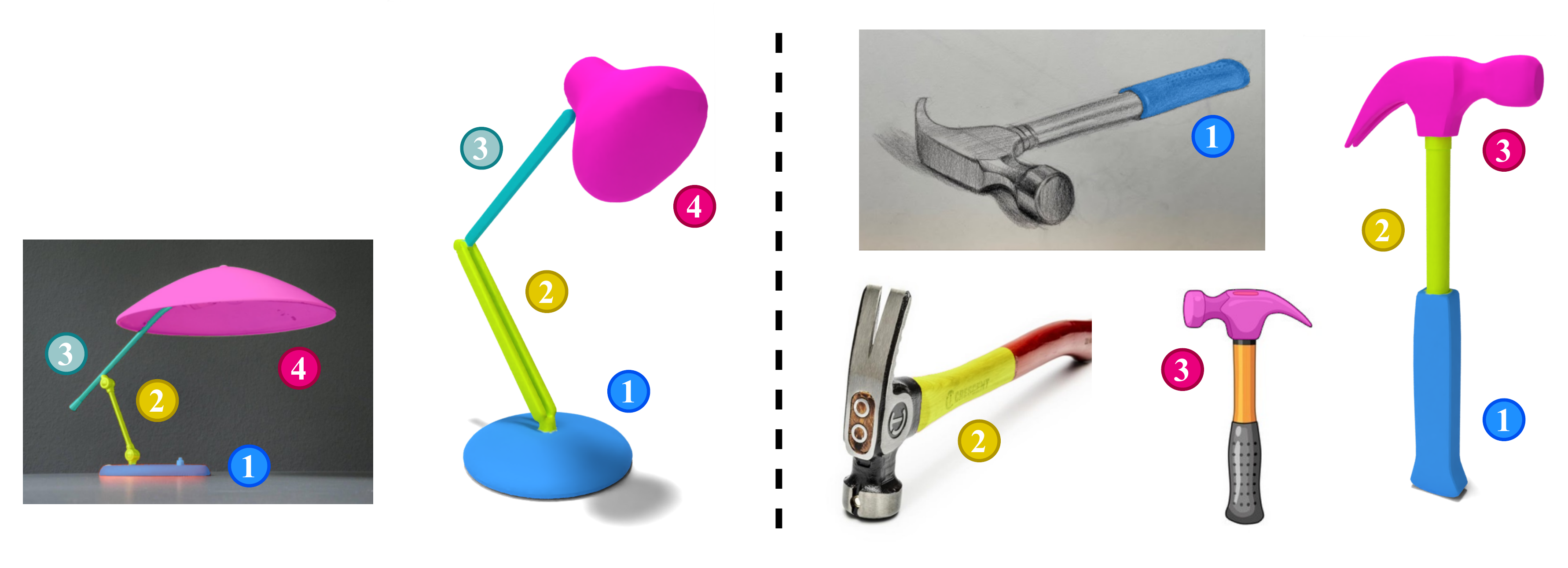}
    \vspace{-7mm}
    \caption{\textbf{Complete segment-to-segment correspondence.} Our method is capable of generating a complete segmentation-to-segmentation correspondence between an image and a shape (left). We can also match corresponding segmentations across a variety of images of different types (sketch, photo, and drawing), poses, and appearances (right).}
    \vspace{-2mm}
    \label{fig:full_shape}
\end{figure}

\subsection{Corresponding Shape Part} \label{sec:shape_seg}
For a pixel that has a best segmentation buddy vertex, we are left with finding the shape part corresponding to the segmented image region. To do so, we leverage a recent work for 3D segmentation based on vertex clicks \cite{lang2024iseg}. iSeg \cite{lang2024iseg} distills a 2D segmentation model and trains a network for predicting the 3D part of a mesh for a given vertex click. This 3D segmentation model predicts a mask $M^{3D}_{v}$ containing the mesh vertices corresponding to the vertex click $v$. In our work, we use the best segmentation buddy $v_p$ to segment the mesh. The resulting region $M^{3D}_{v_p}$ is regarded as the matching 3D part for the 2D segment $M^{2D}_p$ in the image, yielding the desired image-shape correspondence.

\section{Experiments} \label{sec:experiments}

We evaluated our image-shape correspondence method (\ourmethod{}) across a variety of axes. First, \cref{sec:generality} showcases the method's generality. Then, we discuss the method's fidelity in \cref{sec:generality}. Finally,~\cref{sec:robustness} demonstrates the robustness of \ourmethod{} in terms of the considered images and shapes. Additional experiments appear in the supplementary material.

We apply our method to images in the wild, paired with diverse meshes from different sources: COSEG \cite{coseg_2011}, PartNet \cite{mo2019partnet}, Turbo Squid \cite{turbosquid}, Thingi10K \cite{Thingi10K}, Toys4k \cite{Toys4k}, and SHREC '19 \cite{melzi2019shrec}. \ourmethod{} is highly flexible and operates on images and shapes from different categories, such as animals, humanoids, vehicles, and household objects. 

\begin{figure}[!t]
    \centering
    \includegraphics[width=\linewidth, trim=0 0 0 0, clip]{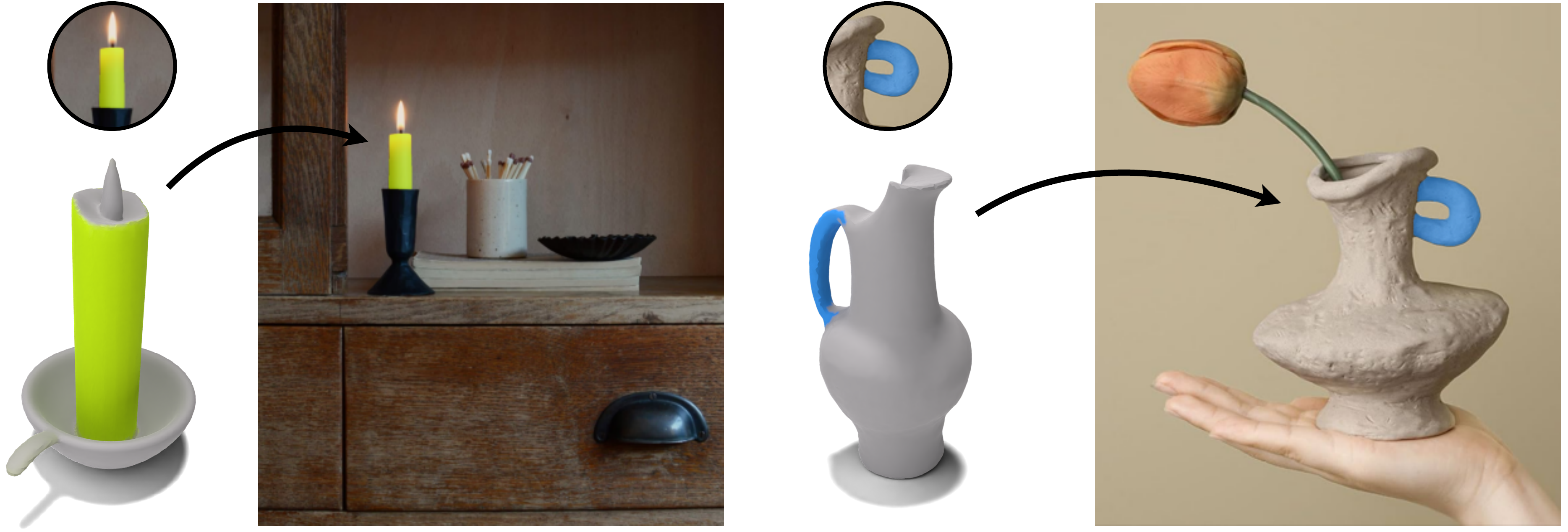}
    \caption{\textbf{Shape to image correspondence.} \ourmethod{} is highly flexible and operates in both directions. In addition to matching an image segment to a 3D part, it can also match a 3D segmentation to the corresponding semantic image region.}
    \vspace{-2mm}
    \label{fig:sh_to_im}
\end{figure}

\ourmethod{} is implemented in PyTorch~\cite{pytorch}. We use DINOv2~\cite{dinov2} as the 2D vision model for finding the best segmentation buddies between the image and the shape, and we also show that our method can work with a different 2D feature-extraction model for correspondence \cite{tang2023imagediffusion}. We take the $k=100$ most similar vertices in the feature space to the pixel $p$ as candidates for matching. An ablation experiment for the configuration of $k$ is provided in the supplementary.

For 2D segmentation, we utilize SAM \cite{kirillov2023segment}, and iSeg \cite{lang2024iseg} is employed for segmenting 3D meshes. We have also experimented with SAM2 \cite{ravi2024sam2segmentimages} as the backbone model for segmentation and did not notice a substantial difference in the image-shape correspondence results. Thus, since iSeg used SAM, we use SAM as our 2D segmentation backbone as well. Inferring the image-shape correspondence for an image click takes about 4 seconds on an Nvidia A40 GPU.

\subsection{Generality of \ourmethod{}} \label{sec:generality}
Our method is highly versatile and handles a wide range of image-shape pairs, as demonstrated in \cref{fig:teaser,fig:gallery,fig:full_shape}. For example, \cref{fig:teaser} shows the correspondence of a hammer's head between an image and a mesh, while the part differs substantially in \textit{shape and appearance} across the modalities. \cref{fig:full_shape} further illustrates the correspondence between a complete segmentation of a 3D lamp object and an image of a lamp, where the geometry of each of the matched parts is considerably different, \eg, the lamps' shade and base.

Additionally, we can match parts for objects from different \textit{domains} - the owl's body and the airplane's ``body''. \cref{fig:gallery} shows additional challenging cross-domin examples, such as relating the flame of the Liberty statue's torch to the flame of a candle mesh, and matching the tail of Splinter to the 3D elephant's tail. Furthermore, \ourmethod{} is able to match fine-grained parts, such as the Toucan's and penguin's beaks and the fists of the boxing kangaroo and the human.

Although we focus on image to shape correspondence, we note that \ourmethod{} is bidirectional and can also match a part of a 3D mesh to the corresponding semantic region in an image. In this case, the image and the shape switch roles - we click on a vertex of the mesh, segment the 3D shape, find the image pixel that is the best segmentation buddy of the clicked vertex, and segment the image using this pixel. \cref{fig:sh_to_im} shows a couple of such examples.

We note that matching a shape to an image poses distinct and interesting challenges. Most images in the wild contain additional distracting objects, background elements, and textures. For example, the candle image in \cref{fig:sh_to_im} also includes a pencil holder and a bowl on top of the dresser drawers. Nonetheless, our method manages to locate the correct part of the correct object in the image.

We attribute the ability to handle the difficult image-shape (and shape-image) correspondence problem to our proposed \ourmethod{} mechanism. When distilling the 2D vision model to 3D, the image and mesh do not match at the pixel-vertex level. This is due to the domain gap manifested as differences in appearance, geometry, and viewpoint.

To overcome this mismatch, our technique finds a vertex in a region of the mesh that corresponds to the image \textit{region} segmented by the pixel click. We note that the nearest neighbor pixel of the vertex in the feature similarity sense is typically not the clicked pixel (see \cref{fig:system,fig:best_seg_buds}). Instead, it is required to fall within the segmented area in the image. This relaxed condition accommodates the feature discrepancy between the modalities and results in segment-to-segment correspondence between the image and the shape.

\subsection{Fidelity of \ourmethod{}} \label{sec:fidelity}

We find correspondences between in-the-wild images and 3D shapes, and we do not have ground-truth annotation of such matches. Thus, as a proxy for the match fidelity, we compute the Intersection over Union (IoU) between the image segment obtained by the pixel click $p$ and the pixel $q^*$ corresponding to the vertex $v_p$ used for the 3D mesh segmentation (see \cref{eq:2D_IoU,eq:max_IoU_pix}).

\begin{table}[t!]
\centering
\begin{tabular}{lccc}
\toprule
Method & NBB \cite{aberman2018neural} & DIFT \cite{tang2023imagediffusion} & \ourmethod{} (ours) \\
\midrule
Success rate $\uparrow$ & 0.64/0.66  & 0.39/0.48 & \textbf{0.74}  \\
\bottomrule
\end{tabular}
\vspace{-2pt}
\caption{\textbf{Quantitative evaluation.} We compare the image-shape correspondence success rate of different zero-shot correspondence techniques on shapes from the PartNet dataset \cite{mo2019partnet} and images generated based on their renderings. For the image-based methods \cite{aberman2018neural, tang2023imagediffusion}, we report the accuracy when using different/the same views of the shape. \ourmethod{}'s matching success rate is better than that of the alternatives.
}
\vspace{-2mm}
\label{tab:quantitative_evaluation}
\end{table}

Interestingly, we have found out that this quantity reflects whether the correspondence between an image region and a 3D mesh part exists. \cref{fig:bsb_vs_non_bsb} exemplifies this phenomenon. For example, the guitar head in the image has a match in the guitar mesh. In this case, the nearest neighbor pixel $q^*$ of vertex $v_p$ is inside the segmentation region of $p$ and results in a segmentation region very similar to that of $p$.

However, for the volume knob region, a best segmentation buddy does not exist. In this case, for each of the vertices $\{v'\}$, the nearest neighbor pixel $q'$ falls \textit{outside} the knob region and yields a significantly different segmentation with almost no intersection with the knob area. To evaluate this property quantitatively, we sampled 100 pixels uniformly from the guitar object image in \cref{fig:bsb_vs_non_bsb}. We have found out that the average IoU for pixels within a region that has a matching 3D part was $0.98$ \vs $0.01$ for pixels at texture elements without a corresponding 3D part, suggesting a strong relation between the low IoU and absence of correspondence, as well as the fidelity to the queried image part in case a matching 3D part exists.

\begin{figure}[!b]
    \centering
    \includegraphics[width=\linewidth, trim=0 70 40 75, clip]{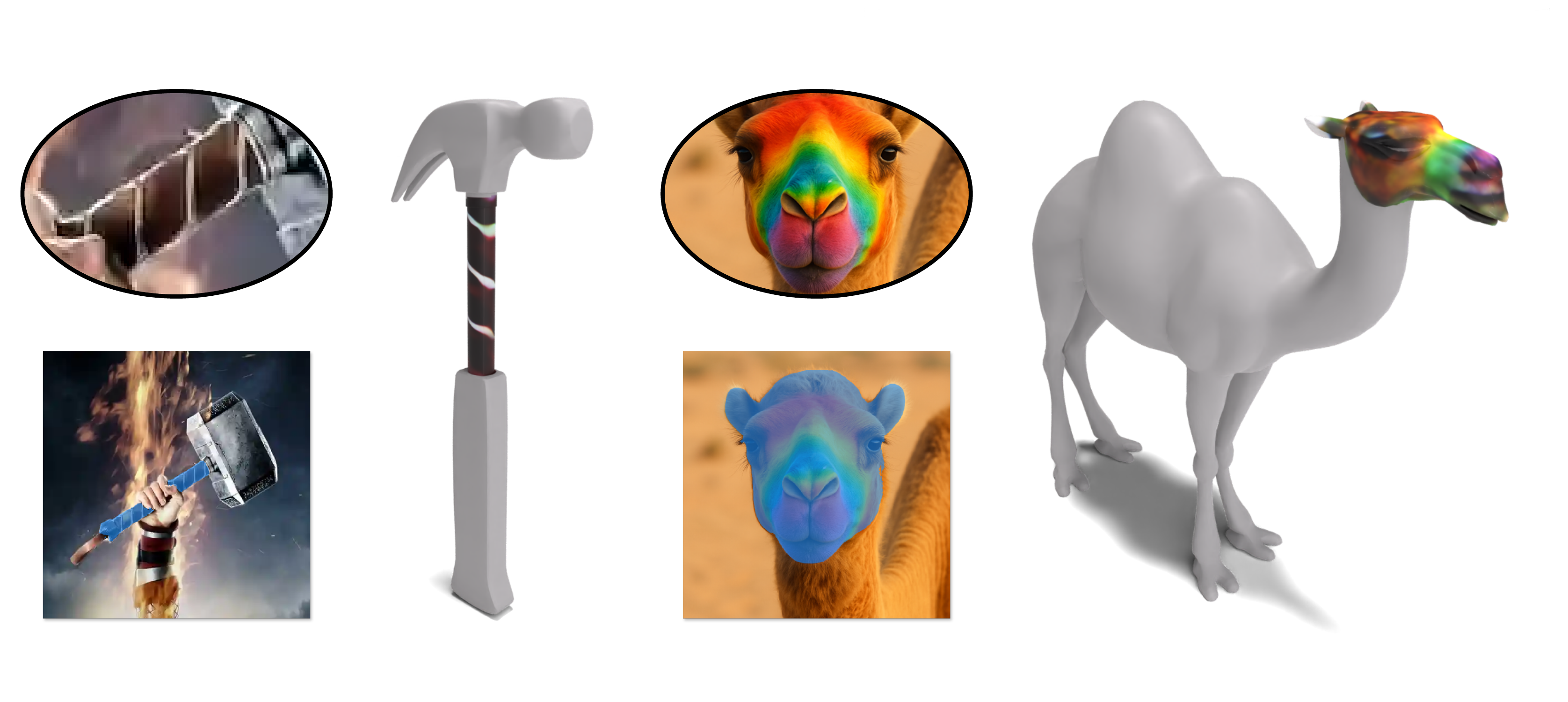}
    \vspace{-5mm}
    \caption{\textbf{Local texturing.} Our image-to-shape matching enables automatic, localized texturing of the shape driven by the texture in the image.}
    \label{fig:local_texturing}
\end{figure}

\ourmethod{} exhibits correspondence fidelity for matching different parts of the same shape to images, as demonstrated in \cref{fig:full_shape}. For example, the image of the lamp and the lamp mesh both have two rods, which are similar in their semantic meaning but geometrically distinct. Despite the resemblance, our method aligns the correct rods between the image and the shape. Additionally, the hammer parts are successfully matched across several different images, although the hammer images have significantly different viewpoints and styles (a sketch, a photo, and a drawing). This high matching fidelity enables finding correspondences for a complete segmentation of the mesh.

\smallskip
\noindent \textbf{Quantitative evaluation.} As far as we can ascertain, there is no annotated dataset for cross-modality image-shape segment correspondence. Thus, inspired by iSeg \cite{lang2024iseg}, we adapted the part segmentation dataset PartNet \cite{mo2019partnet} for our setting. The evaluation included 265 meshes sourced from all the categories in the dataset. For each mesh, we sampled several vertices at random and rendered different views of the shape where the vertex is visible. Then, we used ControlNet \cite{zhang2023adding} to generate colored images conditioned on the depth information from the rendered view, and projected the 3D vertex to a 2D pixel in the generated image. This pixel was regarded as a click on the image, and we evaluated the correspondence success rate - whether our method maps the pixel to a vertex in the ground-truth 3D part to which the original vertex belongs.

As there are no training- and template-free image-to-shape matching methods, we adapted two image-based zero-shot correspondence techniques as baselines: the pioneering work NBB \cite{aberman2018neural} and the recent work DIFT \cite{tang2023imagediffusion}. Existing image-shape correspondence techniques require training supervision \cite{neverova2020continuous,perla2025asia} or are restricted to specific shape templates \cite{neverova2020continuous,shtedritski2024shic}, and thus, are not directly comparable to our method.

\begin{figure}
    \centering
    \newcommand{\pl}{-1.5}
    \begin{overpic}[width=\linewidth, trim=0 -40 70 0, clip]{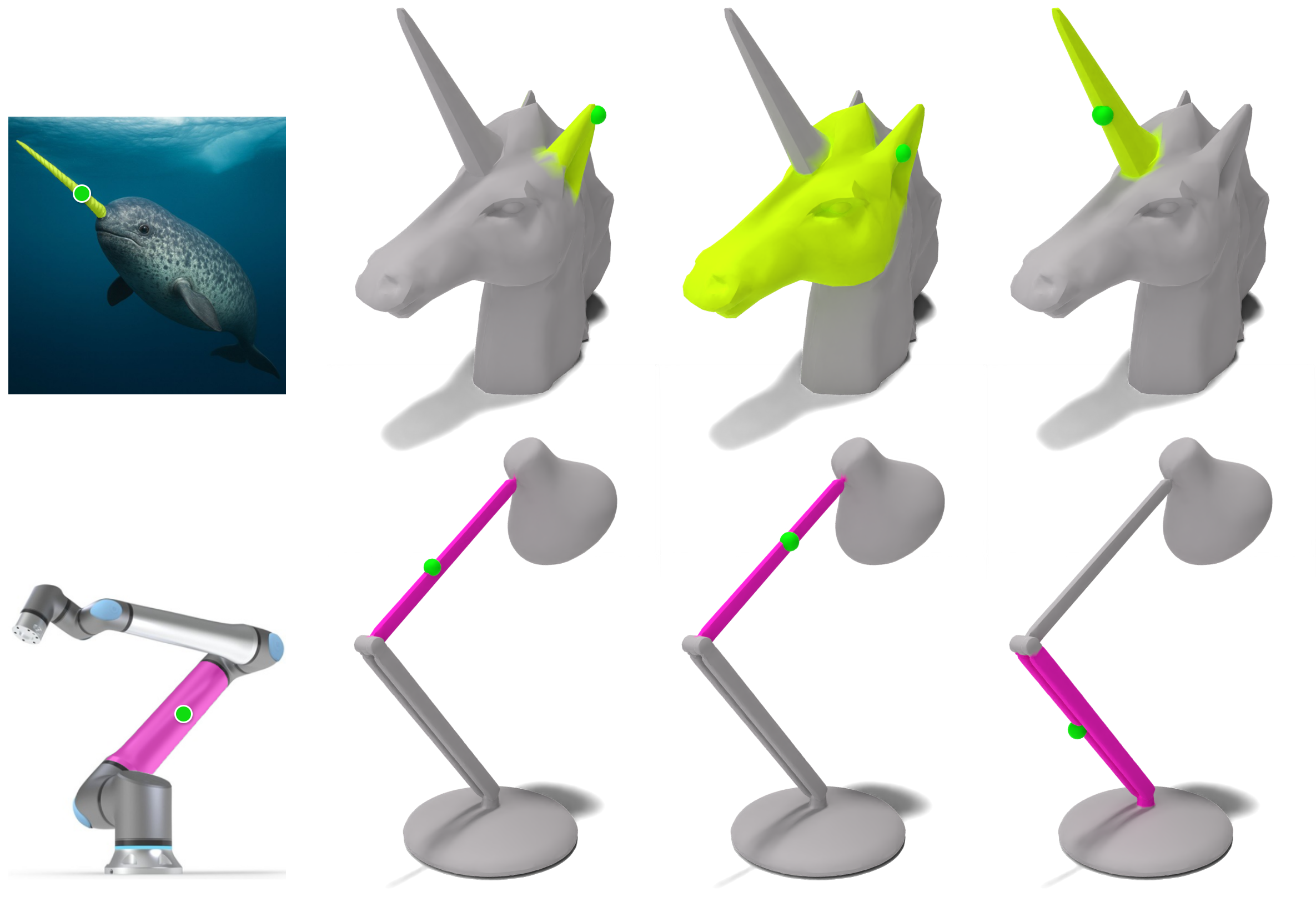}
    \put(31.5, \pl){\textcolor{black}{NBB \cite{aberman2018neural}}}
    \put(56, \pl){\textcolor{black}{DIFT \cite{tang2023imagediffusion}}}
    \put(86.5, \pl){\textcolor{black}{Ours}}
    \end{overpic}
    \caption{\textbf{Qualitative comparison.} We adapt baselines to solve our task from existing techniques \cite{aberman2018neural,tang2023imagediffusion}. These methods produce incorrect correspondences, whereas \ourmethod{} reliably selects the shape part that semantically matches the target image segment.}
    \vspace{-2mm}
    \label{fig:comparison}
\end{figure}

We used the generated RGB images and the renderings of the textureless mesh to evaluate NBB and DIFT. Two settings were considered: matching the clicked pixel from the generated image to the rendered image of the same view or to a rendering from a different view, where the projected clicked 3D vertex is visible. For the evaluation, we obtained the matched pixel in the rendered image, unprojected it back to 3D, and checked whether it fell within the ground-truth part.

\cref{tab:quantitative_evaluation} presents the matching success rate averaged over the evaluation image-shape pairs. NBB relies on a sparse set of mutual nearest neighbor pixels in the neural feature space. Due to the modality gap, these matches are inaccurate, leading to correspondence errors. DIFT computes correspondences using similarity between diffusion model features. This measurement is sensitive to the drastic appearance difference between the textured image and the rendering of the untextured mesh, resulting in wrong alignments. In contrast, \ourmethod{} computes the correspondence between the image and shape directly, and verifies that the vertex is mapped back to the segmented region in the image, yielding correspondences with a higher success rate.

In the supplementary, we further report the results of a perceptual study comparing our method and the baselines for corresponding diverse images and shapes, for which ground-truth annotations do not exist. \cref{fig:comparison} presents such examples. The study showcases that our method is consistently preferred over the alternatives. Similar to the quantitative evaluation, NBB and DIFT were applied to the image and a rendering of the shape. Then, the matched pixel in the rendered image was unprojected to segment the 3D shape.

\begin{figure}[!t]
    \centering
    \includegraphics[width=0.99\linewidth, trim=0 0 0 0, clip]{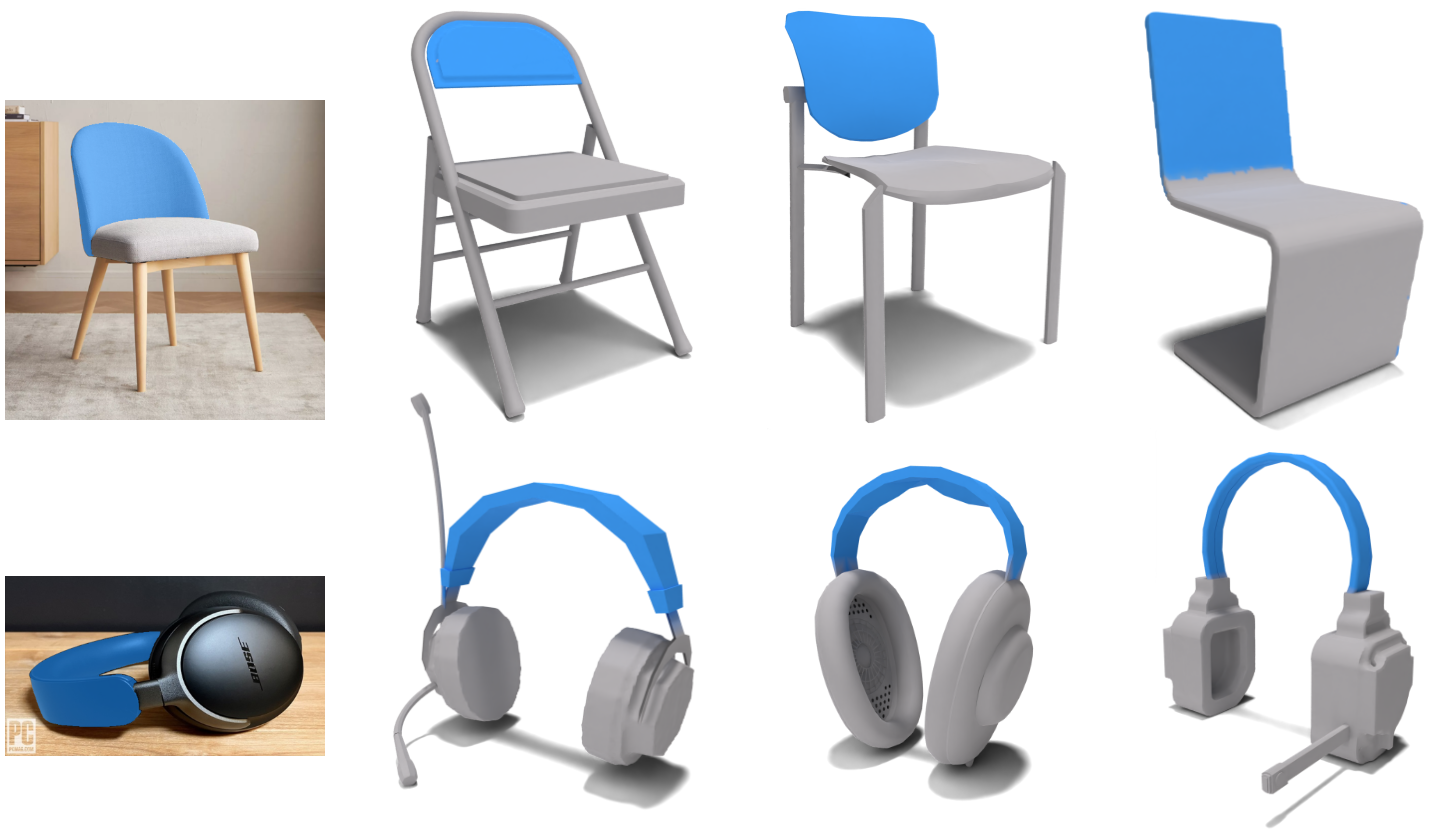}
    \vspace{-3mm}
    \caption{\textbf{Matching the same image to different shapes.} Our method can match regions from images to different shapes that contain significant differences in geometric structures (top) and across occlusions in orientation from the query image (bottom).}
    \vspace{-2mm}
    \label{fig:same_im_diff_sh}
\end{figure}

As seen in \cref{fig:comparison}, the alternative techniques incorrectly match a point on the narwhal's tusk to a point at the unicorn's ear region. These image-based methods are also sensitive to the viewpoint, and wrongly assign the robotic arm's lower rod to the similarly oriented upper rod of the lamp. In contrast, our method finds the correspondence to shape \textit{directly} in 3D and successfully obtains the semantically correct cross-domain correspondences.

\smallskip
\noindent \textbf{Image-guided local shape texturing.} To demonstrate the utility of our correspondence technique, we apply it for controlled mesh texturing. Specifically, we combine our work with an image-guided local 3D texturing method \cite{decatur2025pixbrush}, where the 3D region for texturing is determined by our method. As \cref{fig:local_texturing} shows, our method successfully finds the corresponding 3D part for texturing. 

\subsection{\ourmethod{} Robustness} \label{sec:robustness}
\begin{figure}
    \centering
    \includegraphics[width=0.97\linewidth, trim=0 0 0 0, clip]{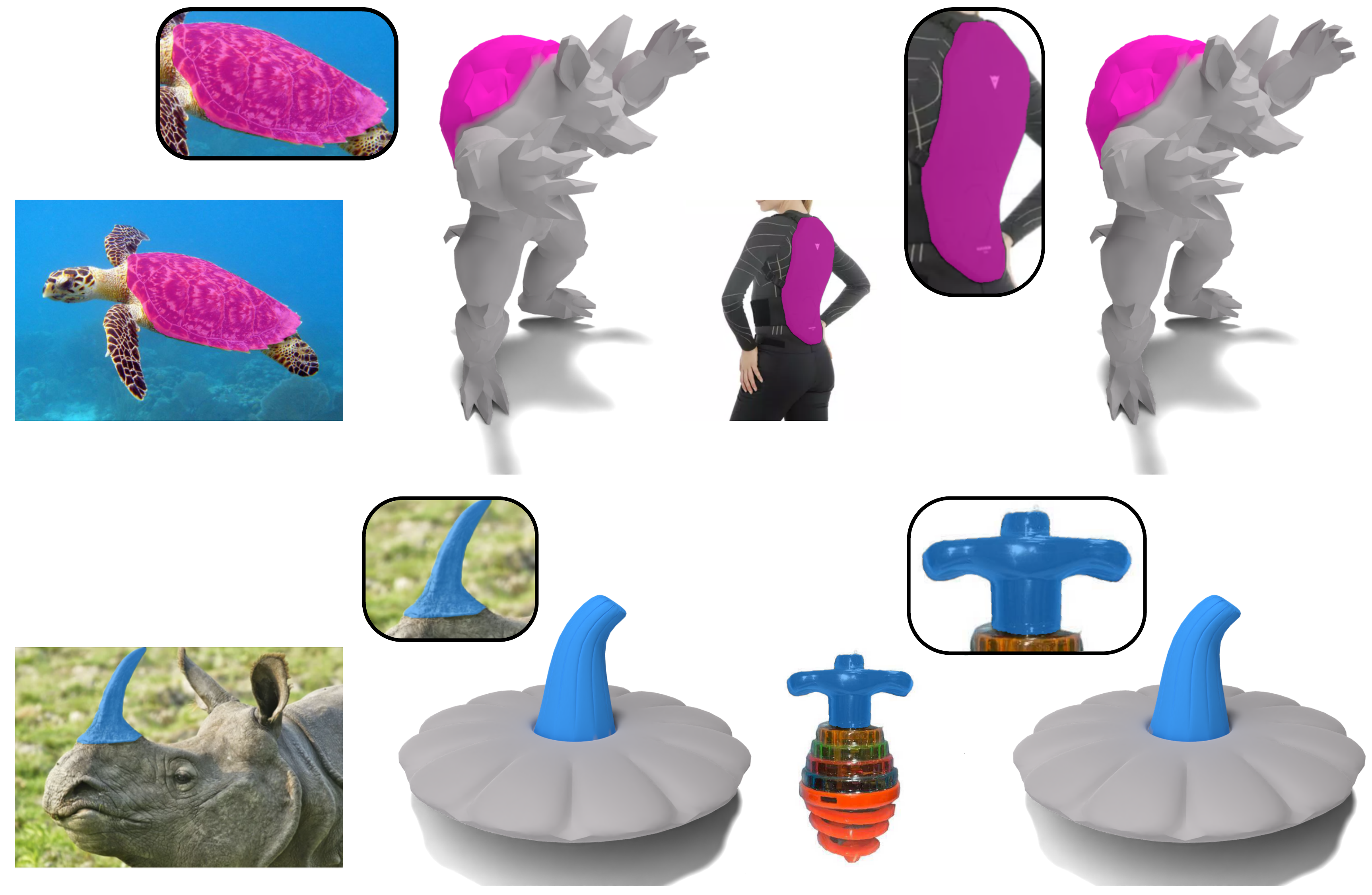}
    \vspace{-2mm}
    \caption{\textbf{Cross-domain image-to-image correspondence.} By matching regions from different images to the same shape part via \ourmethod{}, we indirectly obtain correspondence between different domain images.}
    \vspace{-2mm}
    \label{fig:im_to_im_corr}
\end{figure}

We demonstrate the robustness of \ourmethod{} across different images and shapes. In \cref{fig:same_im_diff_sh}, the same image is matched to different meshes, where the geometry of the overall shape and part differ between the 3D instances. Nonetheless, we accurately segment the relevant part in 3D. An interesting byproduct of our robust correspondence is the ability to find corresponding segmentations across intra-modal examples \textit{without any supervision}, such as the 3D chair backrests or the headphone bridges. \cref{fig:im_to_im_corr} shows the opposite direction, where different images are matched to the same shape, and we obtain unsupervised cross-domain correspondences between image segments, such as between the shell of the sea turtle and the ski back protector.

\section{Conclusion} \label{sec:conclusion}
In this work, we presented \ourmethod{}, a method for finding correspondences across different modalities (image-shape) and domains (\eg owl-airplane). Computing such matches must handle the inherent differences in appearance, geometry, and viewpoint. A key idea that enabled overcoming this gap is relaxing the rigid best buddy constraint to have instead the nearest neighbor of a vertex to lie within a segmentation region of a clicked pixel, such that the matched pixel and vertex are best buddies in the \textit{segmentation} sense.
We demonstrated the effectiveness of our technique on a wide variety of in-the-wild image and shape pairs.

In the future, we are interested in using our segment-to-segment correspondence to influence and update the pretrained visual features to be \emph{correspondence-aware}. An additional natural extension is to leverage our technique to find correspondences across additional modalities, such as videos, 3D Gaussian splats \cite{kerbl20233d}, NeRFs \cite{mildenhall2020nerf}, and more. Finally, another interesting avenue is to explore image and video-driven local deformations of 3D shapes.

\section*{Acknowledgments}

This work was supported by NSF grants \#2402894 and \#2304481, BSF grant \#2022363, gifts from Adobe, Snap, and Google, and the Bennett Family AI + Science Collaborative Research Program. We thank the University of Chicago for providing computational resources and technical support. We also thank the members of the 3DL Lab at the University of Chicago for valuable discussions and feedback, and especially Raj Hansini Khoiwal for her assistance in preparing this work for submission.

{
\small
\bibliographystyle{ieeenat_fullname.bst}
\bibliography{references.bib}
}

\clearpage
\appendix
\maketitlesupplementary

The following sections provide more information regarding our image-shape correspondence method. \cref{sec:bsb_algorithm} refers to the \ourmethod{} algorithm summary. \cref{sec:additional_results} presents additional
results and experiments we conducted with our method. \cref{sec:quantitative_evaluation,sec:perceptual_study} discuss our quantitative evaluation on the PartNet dataset and the perceptual user study, respectively. In \cref{sec:ablation_test}, we detail ablation test results for verifying our design choices. Finally, \cref{sec:implementation_details} elaborates on implementation details for the vision model distillation to the 3D shape and the comparison settings with previous work.

\section{The Best Segmentation Buddies Algorithm} \label{sec:bsb_algorithm}

For clarity, we summarize the \ourmethod{} matching algorithm in \cref{alg:bsb_matching}. 
As explained in the paper, the algorithm includes finding candidate vertices via feature similarity between the clicked image pixel and the mesh vertices, and selecting the vertex whose most similar pixel falls within the part mask and produces the most consistent segmentation with that mask.

\section{Additional Results} \label{sec:additional_results}

\begin{figure}[!t]
    \centering
    \includegraphics[width=\linewidth, trim=0 0 0 0, clip]{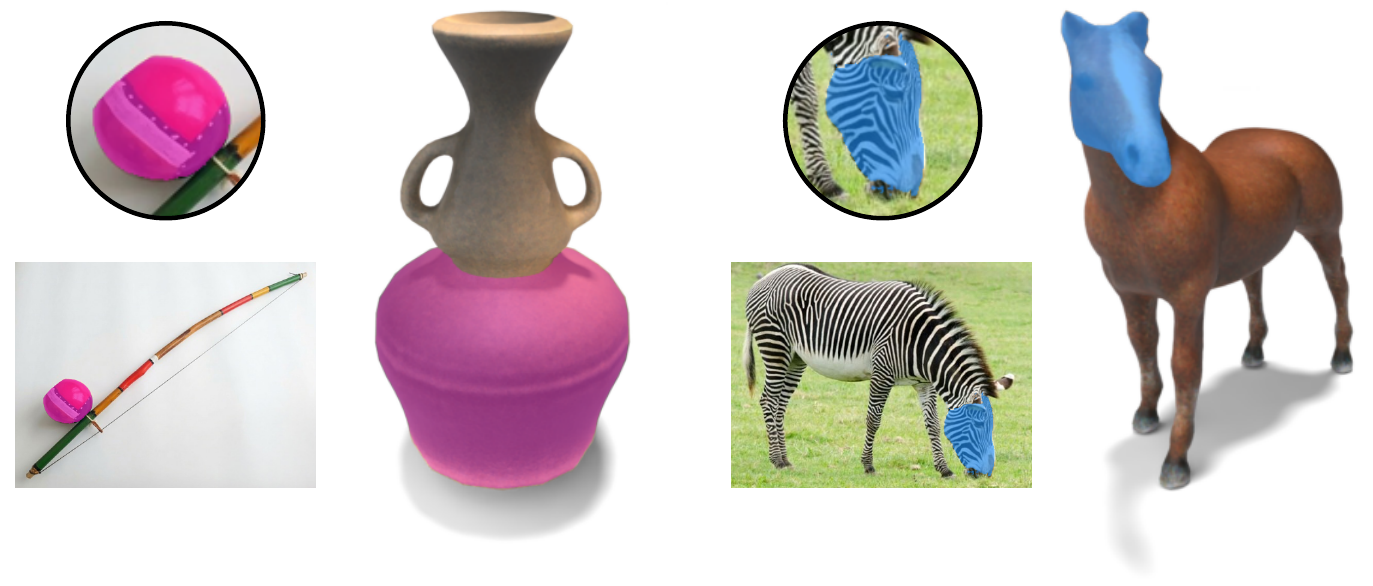}
    \vspace{-8mm}
    \caption{\textbf{Texture robustness.} \ourmethod{} matches semantic regions between image and shape despite variations in their appearance and texture.}
    \label{fig:texture_robustness}
\end{figure}

\noindent \textbf{Robustness to shape texture.} To examine the tolerance of our method to shape texture, we used 3D Paintbrush (without localization) \cite{decatur20223dpaintbrush} to paint untextured meshes, and ran our \ourmethod{} matching scheme, where we distilled 3D vision features from renderings of the textured shape. We found out that our method is robust to the shape’s texture, as exemplified in \cref{fig:texture_robustness}, where the object in the image has a different appearance.

Interestingly, we did not observe a clear advantage in using textured shapes. Our \ourmethod{} mechanism finds a matching vertex in a corresponding semantic region of the shape. Thus, it handles feature discrepancy between the image and the shape and produces robust correspondences when the modalities differ in their texture, and even when the shape is untextured.

\begin{figure}[!b]
    \centering
    \newcommand{\plup}{-4mm}
    \newcommand{\pldn}{-8mm}
    \begin{overpic}[width=\linewidth, trim=0 0 0 0, clip]{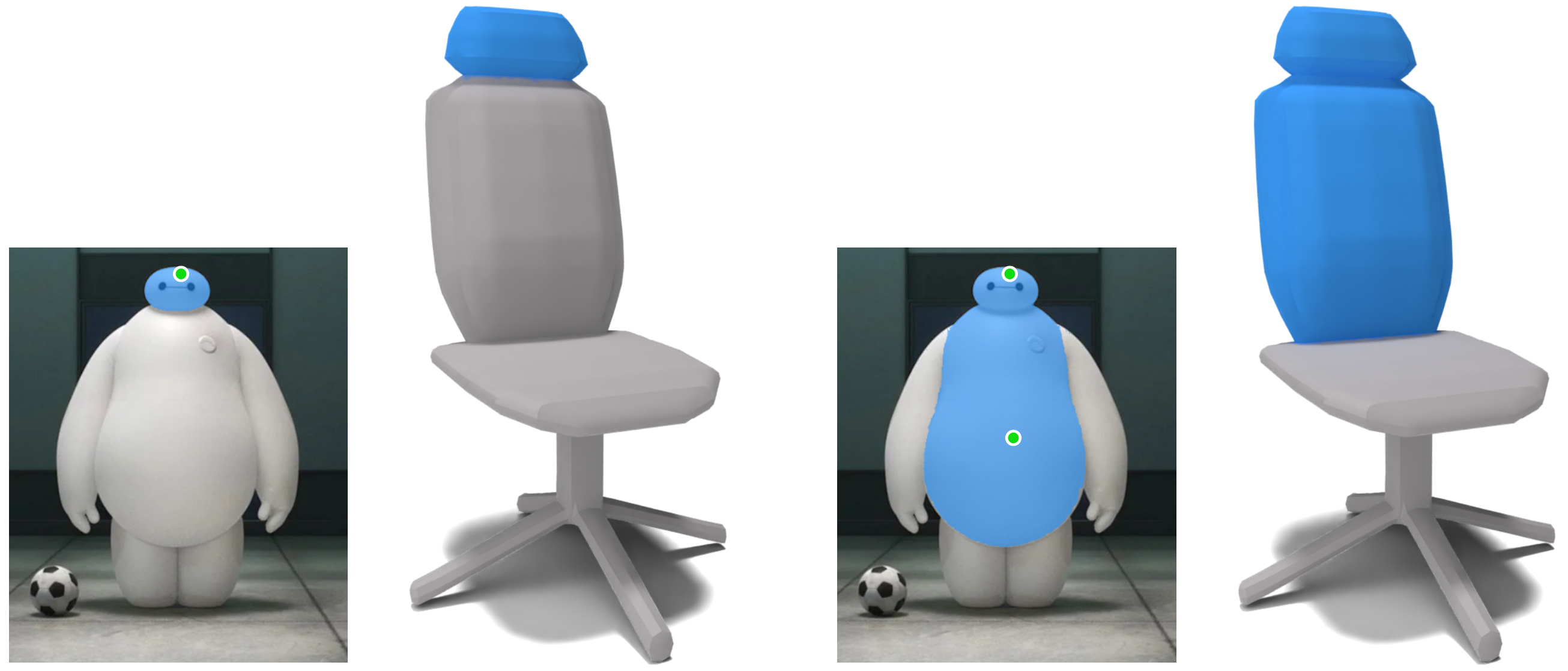}
    \put(8, \plup){\small\textcolor{black}{First}}
    \put(25.5, \plup){\small\textcolor{black}{Corresponding}}
    \put(58.5, \plup){\small\textcolor{black}{Second}}
    \put(78.5, \plup){\small\textcolor{black}{Corresponding}}
    \put(7.8, \pldn){\small\textcolor{black}{click}}
    \put(29, \pldn){\small\textcolor{black}{3D region}}
    \put(60.5, \pldn){\small\textcolor{black}{click}}
    \put(82, \pldn){\small\textcolor{black}{3D region}}
    \end{overpic}
    \vspace{7pt}
    \caption{\textbf{Interactive correspondence.} Our \ourmethod{} matching between pixel clicks and mesh vertices, combined with interactive 2D and 3D segmentation, enables to dynamically update the cross-modality correspondence.}
    \label{fig:interactive_corr}
\end{figure}

\smallskip
\noindent \textbf{Interactive correspondence.}
Since both the 2D and 3D segmentation models are interactive, our \ourmethod{} mechanism enables \textit{interactive} cross-modality cross-domain correspondence, as \cref{fig:interactive_corr} demonstrates. In this experiment, for each clicked pixel, we found the best segmentation buddy vertex. Then, we used the corresponding pixel-vertex pairs successively for the image and the shape, respectively, yielding the interactive edit of the image-shape correspondence.

\begin{figure*}[!t]
    \centering
    \includegraphics[width=\linewidth, trim=0 55 25 20, clip]{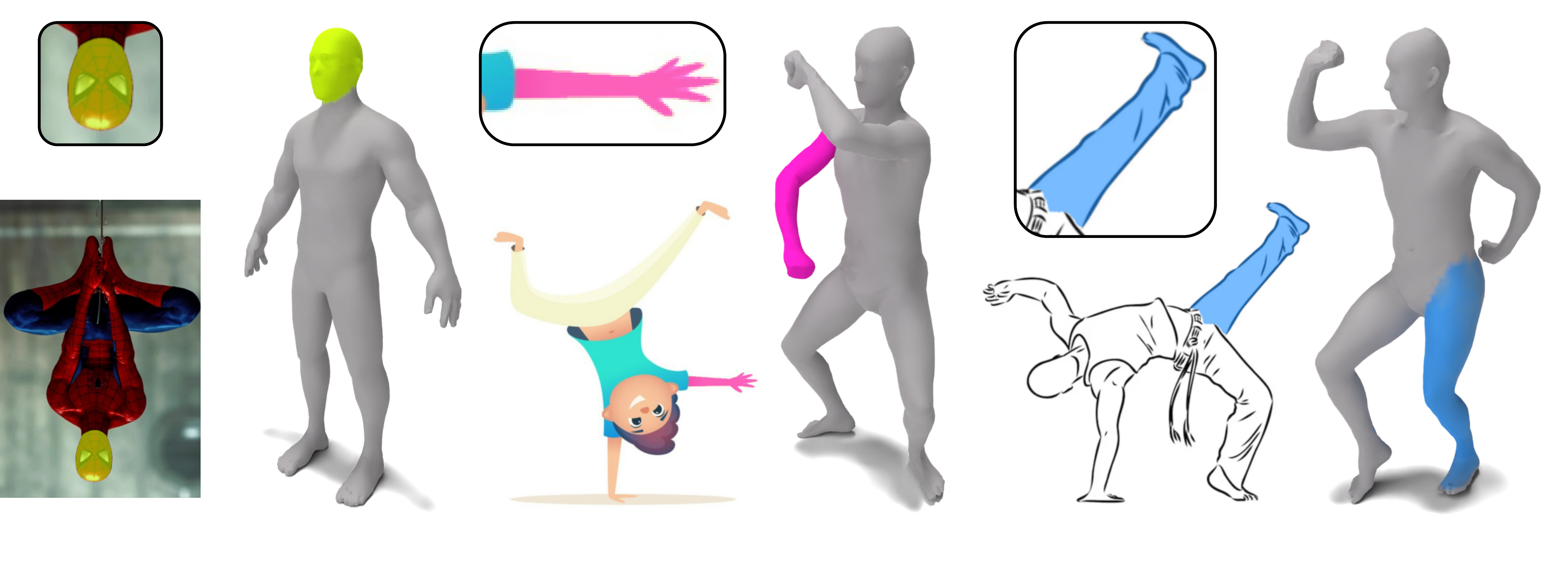}
    \vspace{-7mm}
    \caption{\textbf{Matching between differently posed objects.} \ourmethod{} finds correspondence between the image and shape when the object in each modality differs substantially.}
    \label{fig:different_pose}
\end{figure*}

\smallskip
\noindent \textbf{Different object poses.} Our method can match a region in an image to a 3D shape part when the objects have different poses, as presented in \cref{fig:full_shape} for the rigid hammer shape. In \cref{fig:different_pose}, we show another aspect of this property, demonstrating correspondences for the human body - a deformable object in different poses. As seen in the figure, \ourmethod{} succeeds in matching body parts for extreme pose changes: Spider-Man's head in its famous upside-down pose to an upright A-pose person head; a single hand stand where the right arm is extended to the side to a 3D curved right arm oriented downwards; and the left leg swinging up in a Capoeira cartwheel movement to the downward standing 3D leg. These examples further exemplify the robustness of our correspondences.

\begin{figure}[!b]
    \centering
    \includegraphics[width=\linewidth, trim=0 0 0 0, clip]{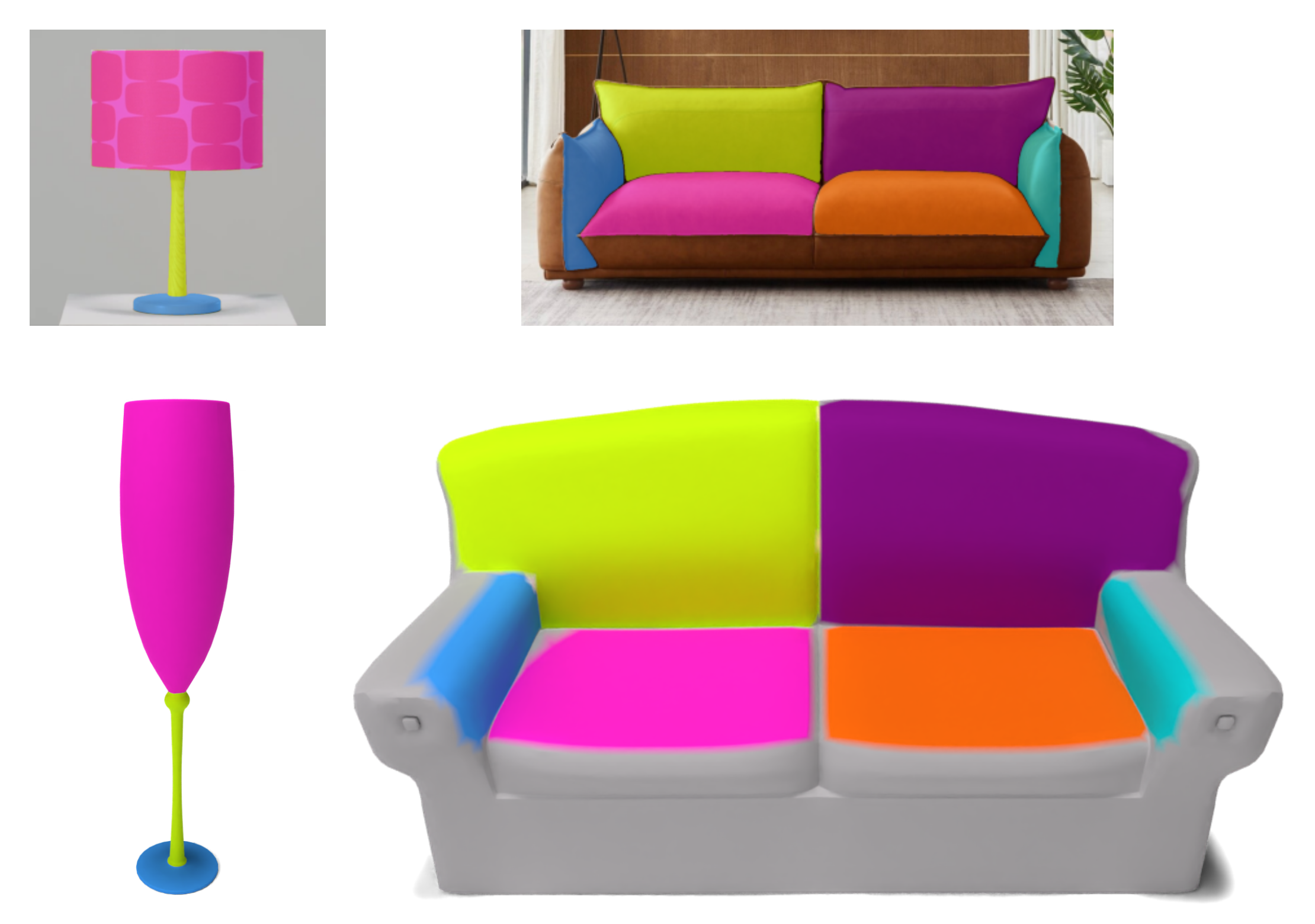}
    \vspace{-5mm}
    \caption{\textbf{Multi-region correspondence.} \ourmethod{} can match multiple regions between the same image-shape pair, when the modalities depict different objects (left), or distinguishing between similar parts of the objects and matching them correctly (right).}
    \label{fig:multi_corr}
\end{figure}

\smallskip
\noindent \textbf{Multi-region correspondence.} In \cref{fig:full_shape} in the paper, we have shown multiple region correspondences between the same image and shape. \cref{fig:multi_corr} shows additional such example. In addition to matching an image and a shape of the same object type (the lamp, \cref{fig:full_shape}), we can also find matches when the overall objects are different, but their parts have a similar semantic meaning (the lamp and goblet pair, \cref{fig:multi_corr}), or when the object has several similar part instances (the backrest and seat pillows and the armrests in the couch pair, \cref{fig:multi_corr}). These results highlight the correspondence specificity of our method.

\begin{figure*}[!t]
    \centering
    \includegraphics[width=\linewidth, trim=0 70 200 220, clip]{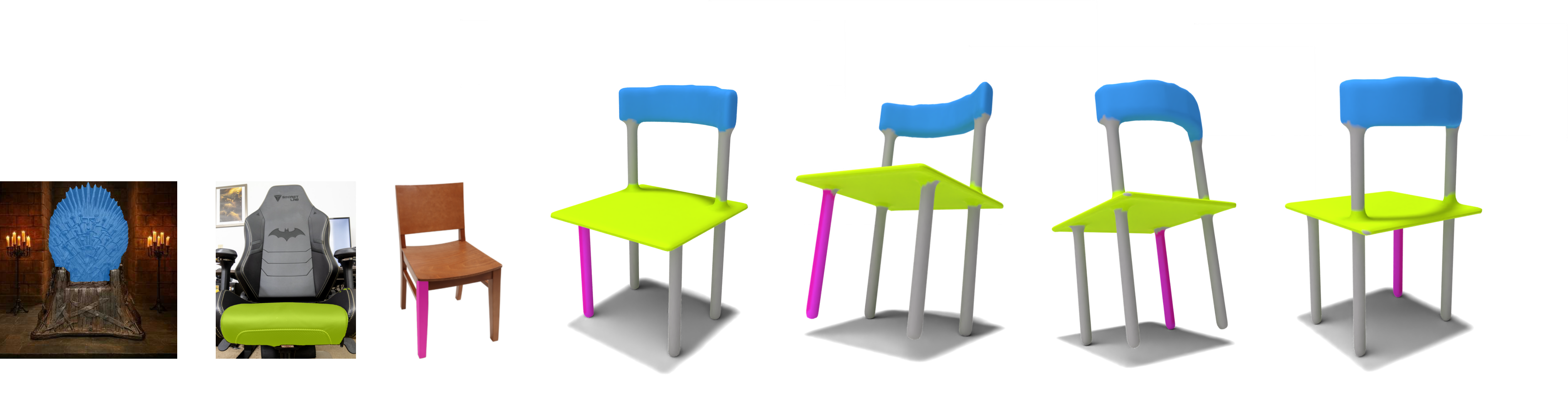}
    \vspace{-6mm}
    \caption{\textbf{A single view to a complete 3D part.} Although each image depicts only one view of the object (left), the entire corresponding part is successfully segmented in 3D (right).}
    \vspace{3mm}
    \label{fig:3d_part}
\end{figure*}

\smallskip
\noindent \textbf{One view to a 3D part.} Our method corresponds a part segment in a 2D image to a complete part in a 3D mesh, as we show in \cref{fig:3d_part}. Surprisingly, even though the image contains only one viewpoint of the underlying object, we are still able to accurately match the entire corresponding region in 3D (see the backrest, seat, and leg in \cref{fig:3d_part}). We hypothesize that this capability can be partially attributed to the fact that we render the mesh from multiple views and lift deep visual features from each of the views onto the shape surface. Thus, a relevant viewpoint of the shape is likely to be present during the distillation process, enabling the matching of a pixel-vertex pair from the corresponding semantic regions.

\begin{figure}[!b]
    \centering
    \includegraphics[width=\linewidth, trim=0 0 70 70, clip]{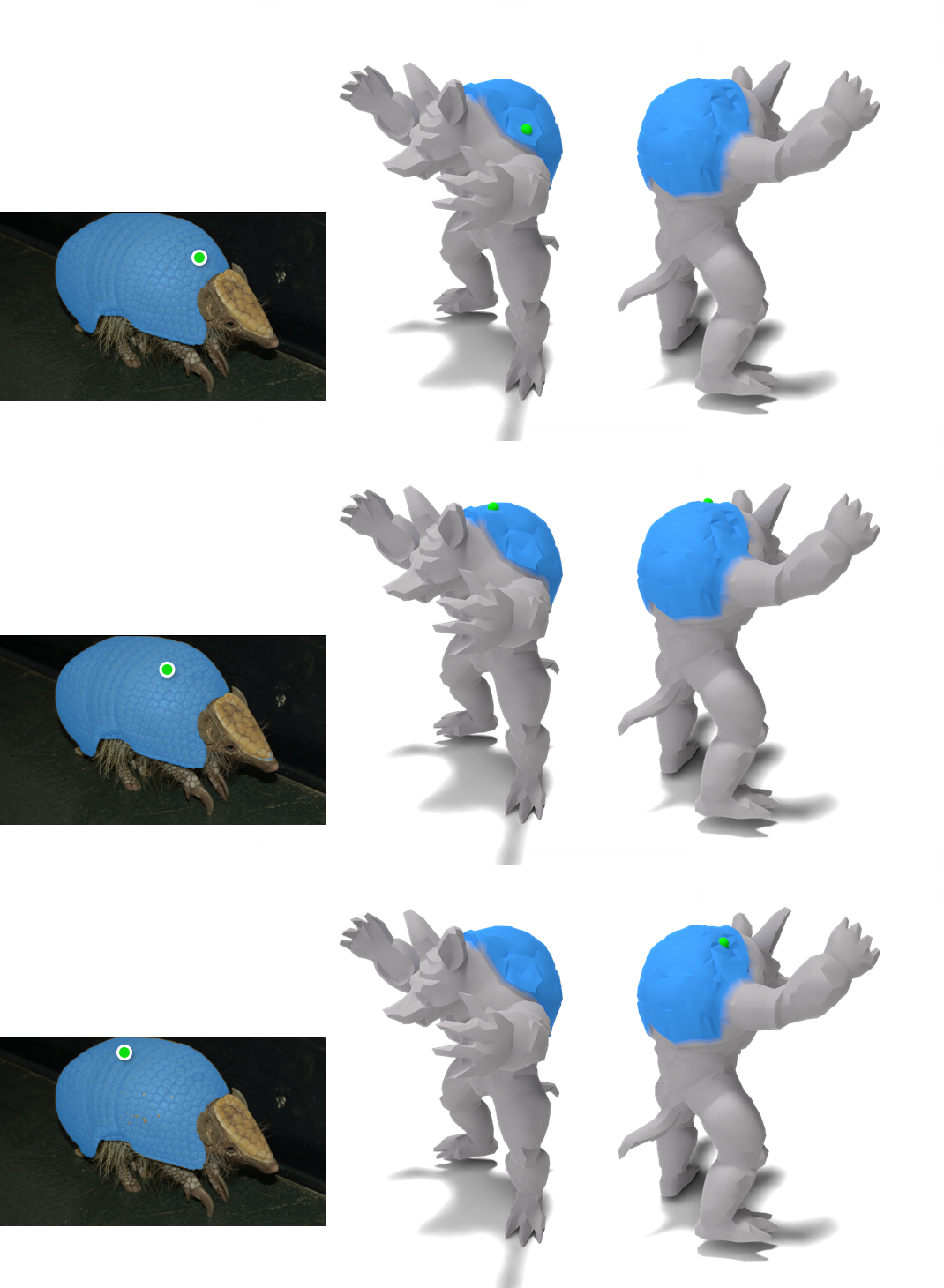}
    \vspace{-7mm}
    \caption{\textbf{Correspondence stability.} Our method is robust to the location of the pixel click in the image region (left). Although different pixels are matched to different vertices, they fall within the corresponding semantic 3D part (right), resulting in a stable matching between the image and the shape. The clicked pixel and the matched vertices are visualized with a green dot.}
    \vspace{-2mm}
    \label{fig:stability}
\end{figure}

\smallskip
\noindent \textbf{Correspondence stability.} \cref{fig:stability} shows the Correspondence results for different clicks within the same image region (the armadillo's carapace). While the clicks are matched to different shape vertices, these vertices fall within the semantically matching region in the standing armadillo shape. Thus, the correspondence between the image and shape is maintained for the different image clicks, demonstrating the stability of our method.

\begin{figure*}
    \centering
    \includegraphics[width=\linewidth, trim=0 0 0 0, clip]{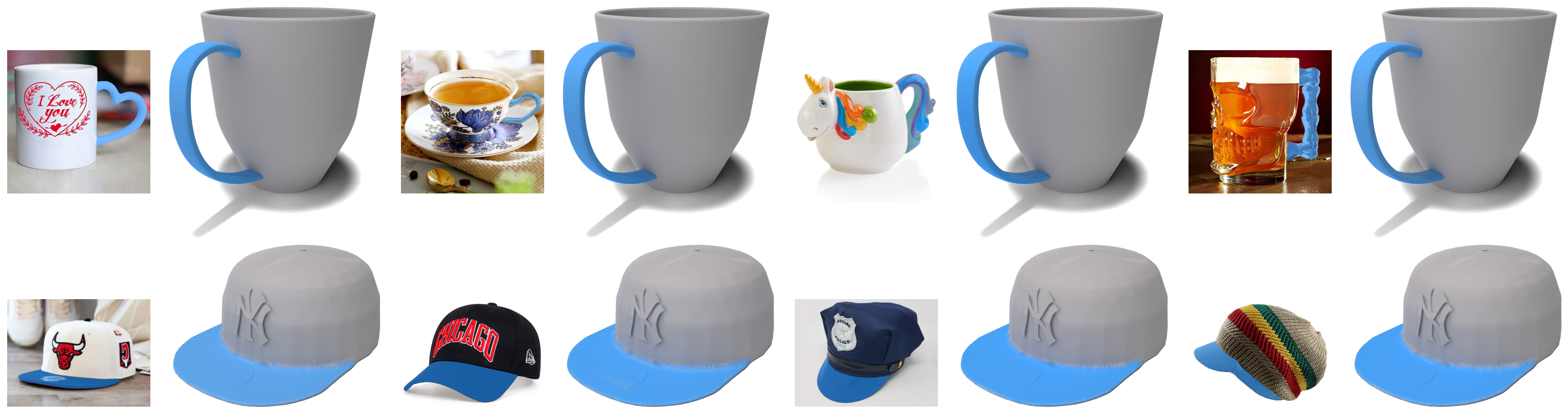}
    \caption{\textbf{Different images to the same shape.} \ourmethod{} accurately matches segmentations from images that contain significant differences in geometry (\eg, the heart-handle on the left) and appearance (\eg, crochet hat on the right) to the same shape. }
    \label{fig:diff_im_same_sh}
\end{figure*}

\smallskip
\noindent \textbf{Corresponding different images to the same shape.}
\cref{fig:diff_im_same_sh} presents the correspondence between several images and the same mug shape. The images depict various types of object instances (a tea cup, a themed mug, and a glass), where the geometry and appearance of the 3D mug handle differ significantly between the images. Nonetheless, our method can consistently segment the corresponding part in 3D, suggesting its robustness to the part properties. Additionally, as explained in \cref{sec:robustness} in the main body, a beneficial outcome of this multi-image to the same shape correspondence is the unsupervised region matching between the images (the cup handles and the hat brims in \cref{fig:diff_im_same_sh}).

\begin{figure}[!b]
    \centering
    \includegraphics[width=\linewidth, trim=0 00 0 0, clip]{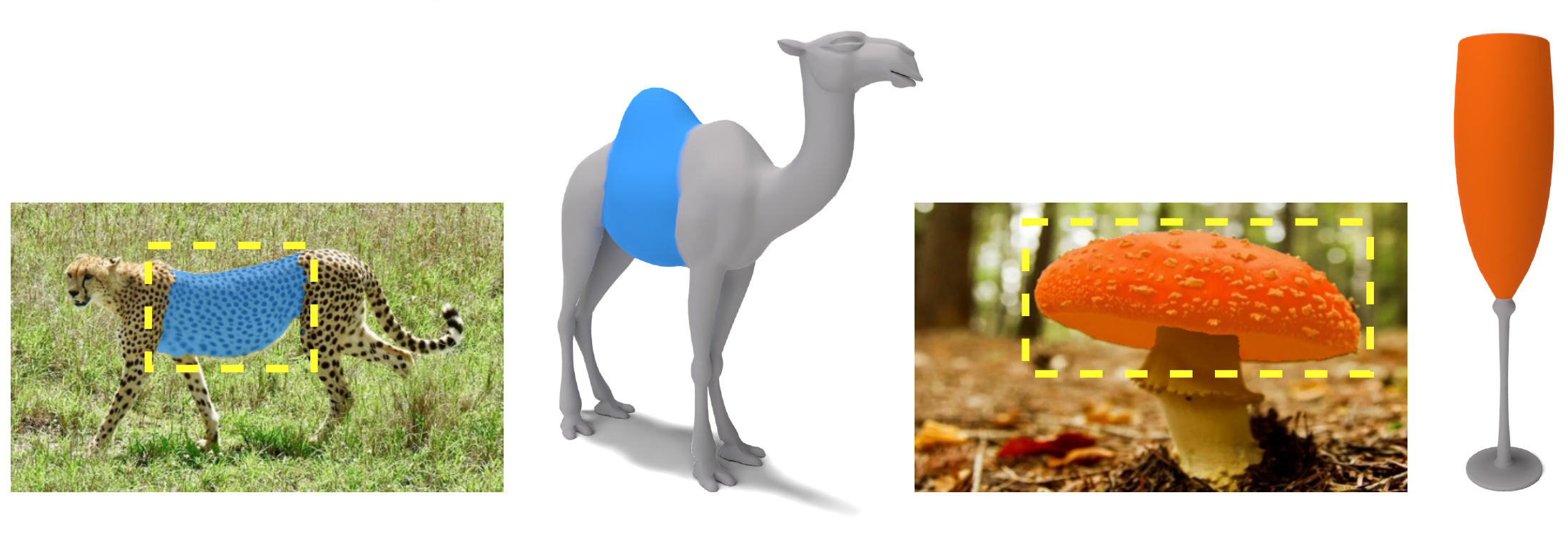}
    \caption{\textbf{2D region segmentation with a box prompt.} \ourmethod{} is highly flexible and can be used with a box prompt (the dashed rectangle) for segmenting the part in the image.}
    \label{fig:box_seg}
\end{figure}

\smallskip
\noindent \textbf{Additional image segmentation interfaces.} Our method is not limited to click-based segmentation. It supports alternative ways for obtaining the part region in the image, in addition to point-click, offering further flexibility to the user. One such approach is to utilize the 2D segmentation model \cite{kirillov2023segment} with a box input, where the user specifies the top-left and bottom-right coordinates in the image to segment the part mask $m^{2D}_p$ used in our matching scheme. Examples are shown in \cref{fig:box_seg}. Another interface for segmenting the image is text, as we describe next.

\begin{figure*}
    \centering
    \includegraphics[width=\linewidth, trim=0 0 0 0, clip]{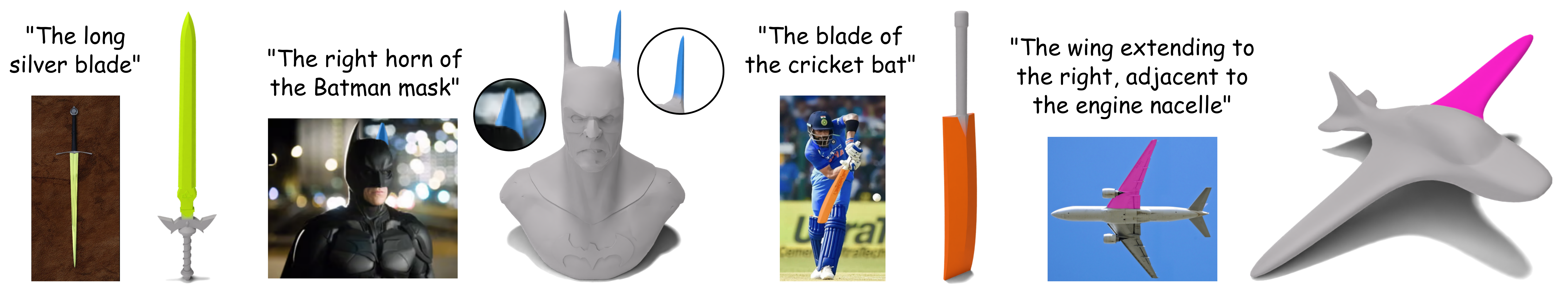}
    \vspace{-5mm}
    \caption{\textbf{Text-based 3D segmentation.} Combining language-driven image segmentation with our method, we achieve 3D segmentation with text. The prompt above the image was used for its segmentation.}
    \vspace{3mm}
    \label{fig:text_seg}
\end{figure*}

\smallskip
\noindent \textbf{Text to 3D segmentation.} In the main paper, we used a click-based model for segmenting the image \cite{kirillov2023segment}. However, another popular interface nowadays for image segmentation is text. Our method can be easily integrated with this interface, as \cref{fig:text_seg} demonstrates. In this experiment, we applied language-driven image segmentation by predicting a bounding box for an object part described by text, and segmenting the part within the bounding box \cite{ravi2024sam2segmentimages}. Then, we used that part mask and its centroid as the pixel click with our \ourmethod{} mechanism to find the corresponding vertex and segment the 3D mesh. This process resulted in text-based 3D segmentation: the image is segmented with text, and the matching 3D part is found by our method.

\begin{figure}[!b]
    \centering
    \includegraphics[width=\linewidth, trim=0 30 0 0, clip]{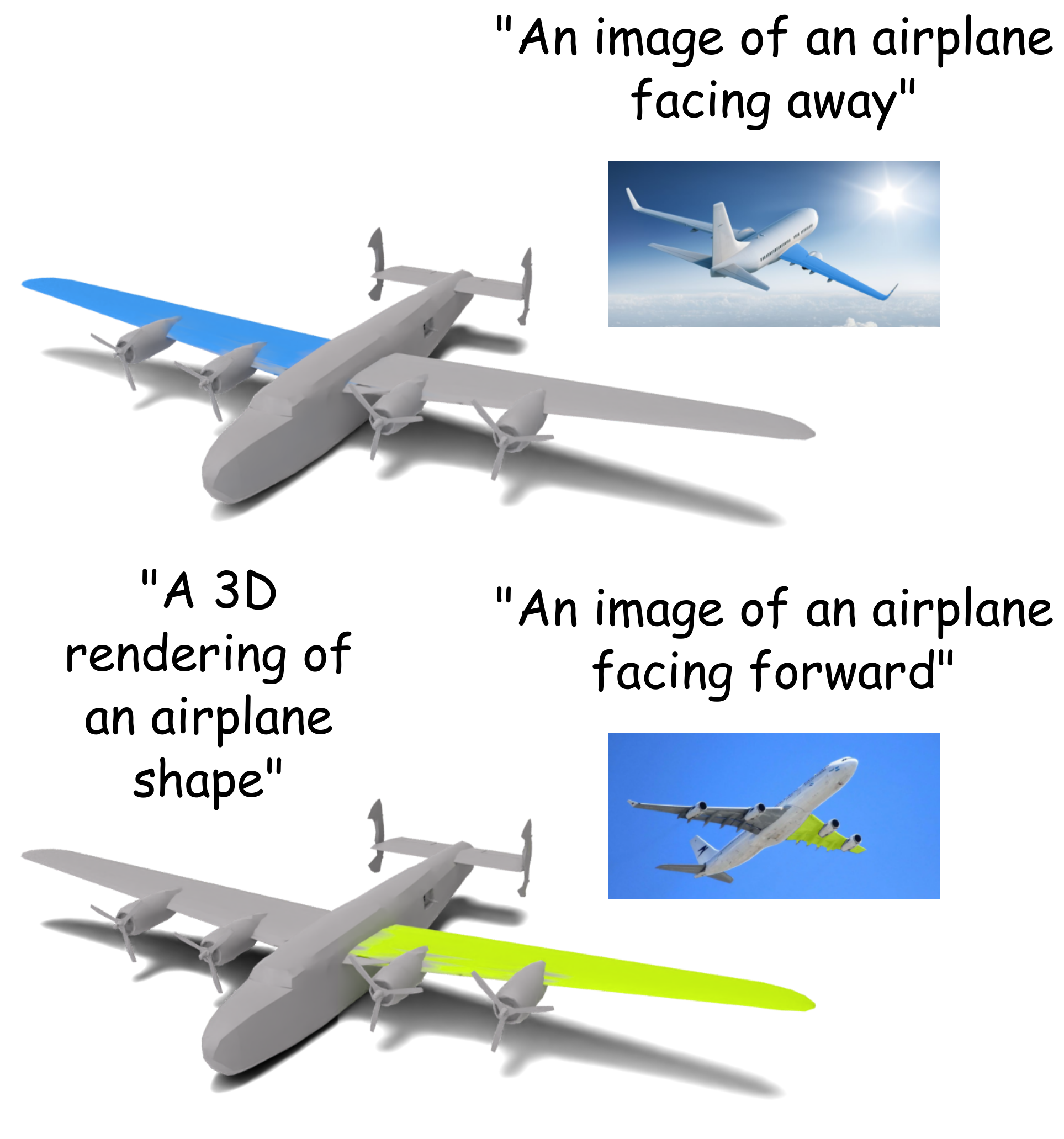}
    \vspace{-6mm}
    \caption{\textbf{Backbone model versatility.} \ourmethod{} can utilize a vision backbone other than DINOv2 \cite{dinov2}. In this case, we lift a diffusion model features \cite{tang2023imagediffusion} to the 3D mesh for finding correspondences. The text prompts used to extract features for the images and the renderings of the shape are indicated next to them.}
    \vspace{-2mm}
    \label{fig:left_right}
\end{figure}

\smallskip
\noindent \textbf{Different vision model backbones.}
Recent works have devised new feature-extraction methods from 2D foundation models for image-to-image correspondence \cite{tang2023imagediffusion, luo2023dhf, zhang2023tale, hedlin2023unsupervised}. Our image-shape correspondence framework is versatile and can incorporate such techniques. \cref{fig:left_right} shows an example where Diffusion Image Features (DIFT) \cite{tang2023imagediffusion} were used for finding best segmentation buddies (instead of DINOv2~ \cite{dinov2}). For example, this backbone vision model enables matching the wings for image-mesh airplanes. However, we note that DIFT requires text prompts describing the image and the shape, while DINOv2 does not require such inputs from the user. Thus, in our experiments, we have focused on using DINOv2 as the backbone model for feature correspondence.

We note that this experiment is different than the one described above. Before, we replaced the \textit{segmentation model} to use text instead of a click interface, and employed DINOv2 \cite{dinov2} as the vision model. Here, we changed the \textit{vision model} \cite{tang2023imagediffusion} for extracting pixel and vertex features for comparing their similarity, and used click-based image segmentation \cite{kirillov2023segment}. Together, these experiments showcase the versatility of our approach, where the different segmentation and vision models can be utilized for diverse and flexible image-shape correspondence.

\begin{figure}[!b]
    \centering
    \includegraphics[width=\linewidth, trim=0 0 0 0, clip]{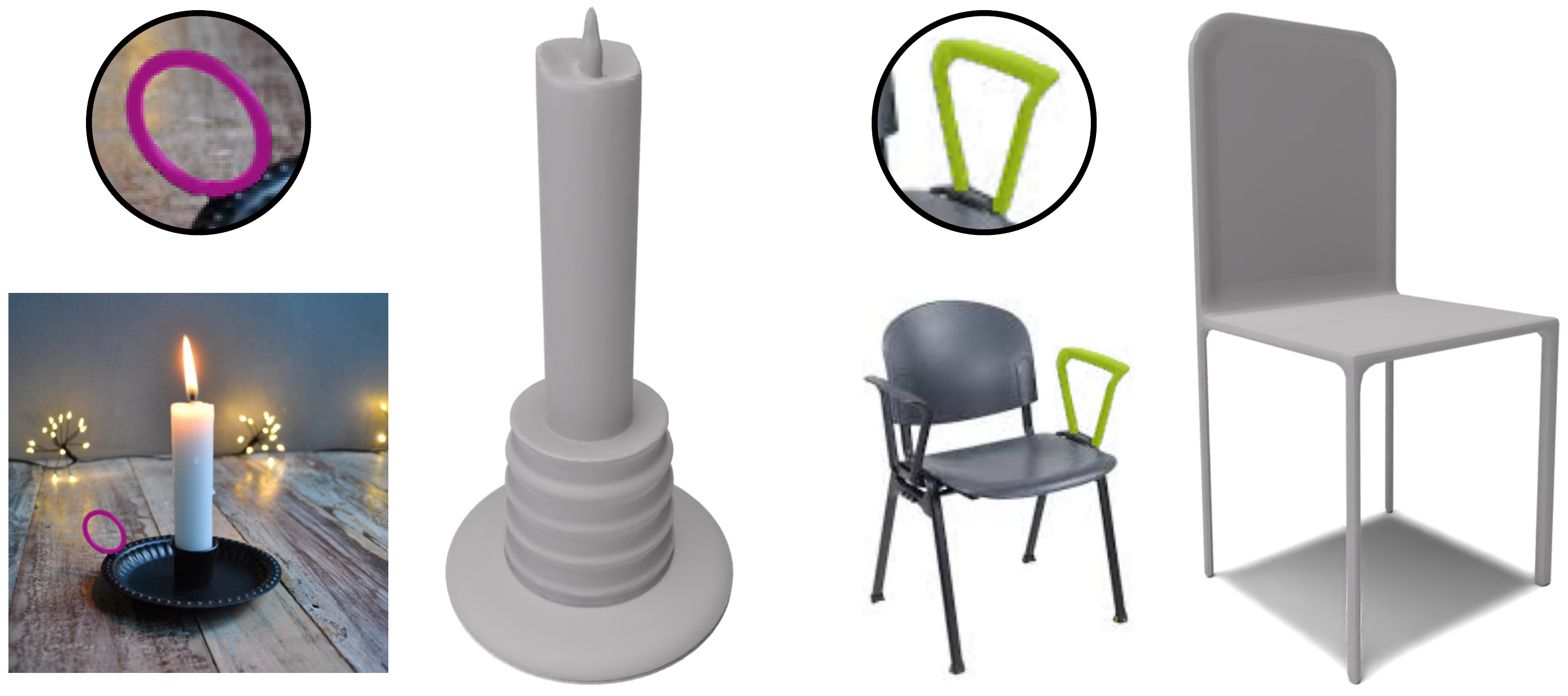}
    \vspace{-5mm}
    \caption{\textbf{Missing shape part.} If a segmented region in the image is missing a matching part in the shape, our method will output an empty 3D segmentation, indicating correctly that correspondence does not exist in this case.}
    \label{fig:missing_part}
\end{figure}

\smallskip
\noindent \textbf{Missing corresponding shape part.}
When the shape is missing a part that exists in the image, our \ourmethod{} scheme will detect that there is no best segmentation buddy vertex, and no part of the shape will be segmented. See examples in \cref{fig:missing_part}.

\begin{figure}[!b]
    \centering
    \includegraphics[width=\linewidth, trim=0 0 0 0, clip]{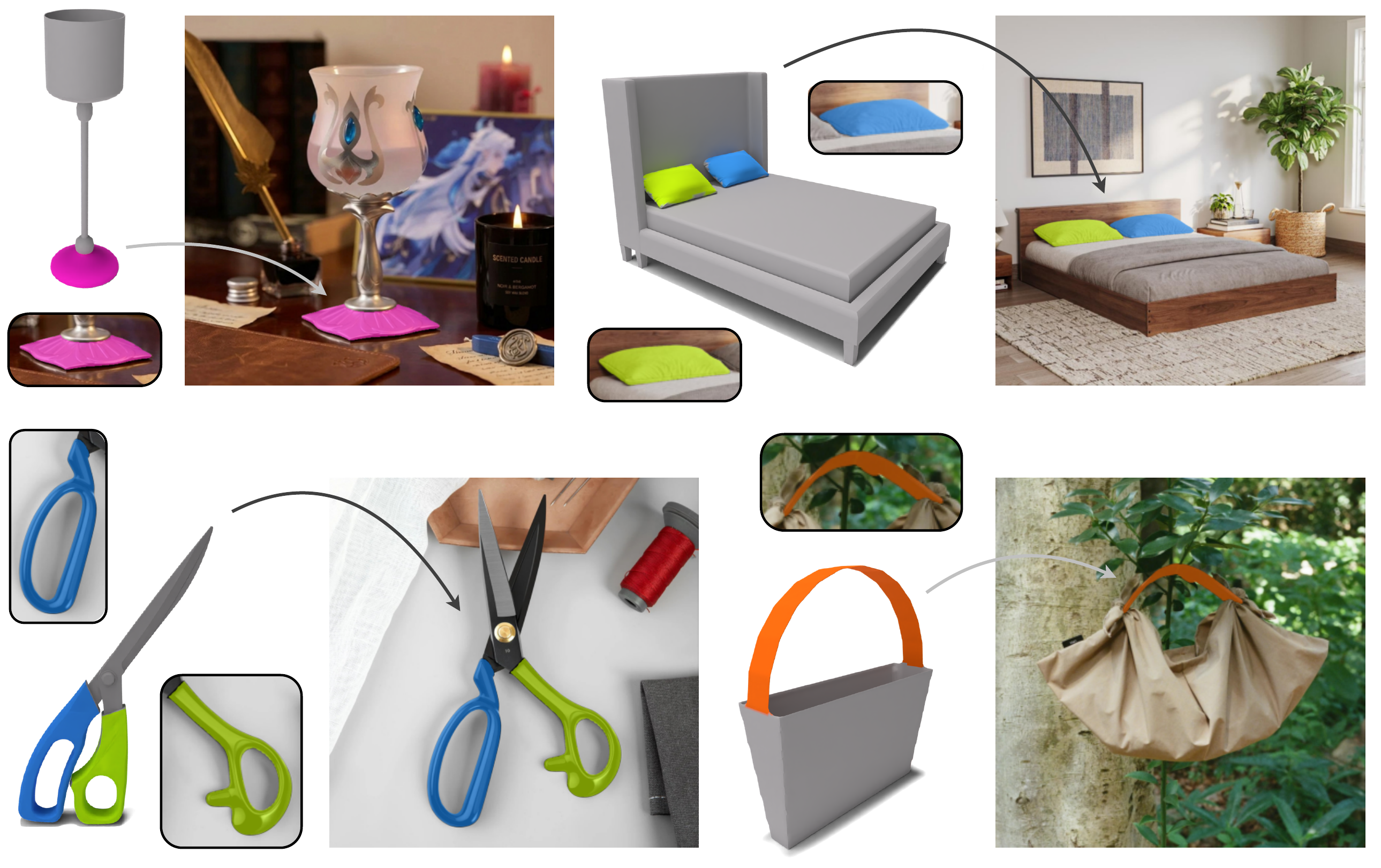}
    \caption{\textbf{Shape-to-image correspondence.} Our method can match a part of the shape to its corresponding semantic segment in the image, even when the image contains a lot of distractors.}
    \label{fig:sh_to_im_supp}
\end{figure}

\smallskip
\noindent \textbf{Shape-to-image correspondence.} As explained in the main paper, \ourmethod{} can be utilized for 3D to 2D matching, when the image and shape switch roles. Additional such examples appear in \cref{fig:sh_to_im_supp}.

\begin{figure}[!b]
    \centering
    \includegraphics[width=\linewidth, trim=0 0 0 0, clip]{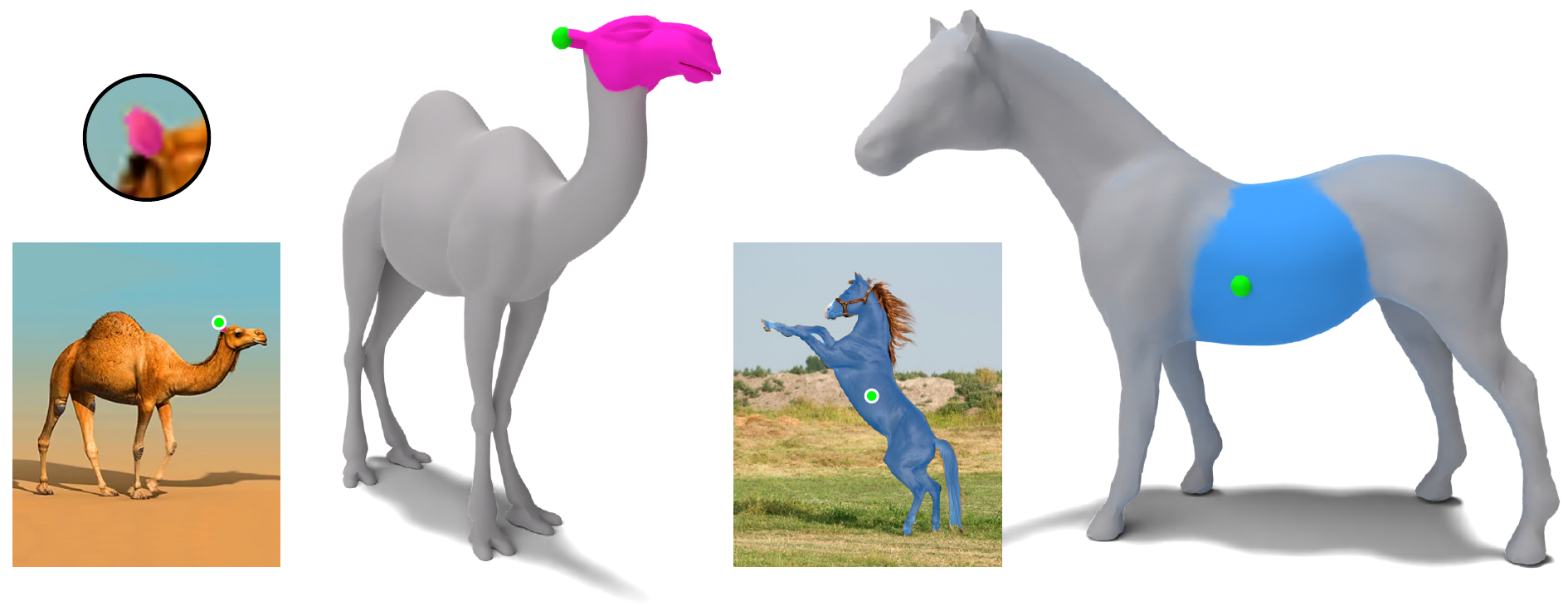}
    \caption{\textbf{Segmentation granularity mismatch limitation.} Our method assumes that the 2D and 3D segmentation models exhibit the same granularity for the image and the mesh. However, correct corresponding clicks on the image and the mesh may have different segmentation mask granularity, compromising the resulting segment correspondence between the modalities.}
    \label{fig:granularity_mismatch}
\end{figure}

\smallskip
\noindent \textbf{Limitations.}
Even though we leverage the same segmentation backbone \cite{kirillov2023segment} to perform segmentation in 2D and in 3D (after distillation), there may be a mismatch in the segmentation \textit{granularity} due to the modality gap. Thus, even if our method finds the best segmentation buddy in the right semantic region of the shape, the 3D and 2D segments may not match. \cref{fig:granularity_mismatch} demonstrates such cases.

For a click on the camel's ear, the 2D model segments the ear. In this case, \ourmethod{} selects the best segmentation buddy vertex correctly on the corresponding ear of the camel shape. However, the 3D model considers the matched vertex to be a part of the head and segments the entire camel's head. In the case of the horse mesh, while the 2D model segments the entire horse body, the 3D model segments only the belly.

\begin{figure}[!b]
    \centering
    \includegraphics[width=\linewidth, trim=0 0 0 0, clip]{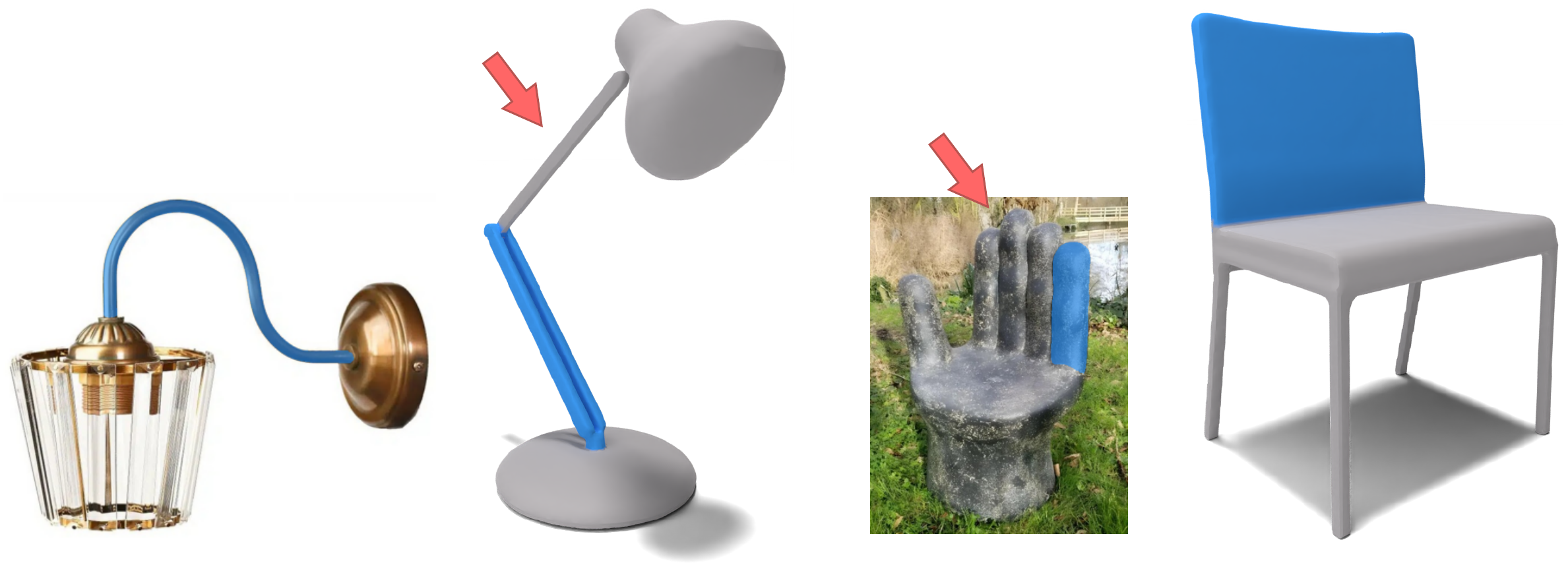}
    \vspace{-5mm}
    \caption{\textbf{One-to-many and many-to-one partial matching limitation.} When a single segment in the image corresponds to multiple mesh parts of the same semantic entity (the lamp's body) or vice versa (the chair's backrest), our resulting image-shape correspondence will be partial.}
    \label{fig:partial_match}
\end{figure}

Another limitation arises when one segment in the image corresponds to several parts in the shape or the other way around, as presented in \cref{fig:partial_match}. In the first case, the curved lamp rod in the image matches the two 3D lamp rods. However, since our method selects only one \ourmethod{} vertex for a pixel click, and the 3D segmentation model \cite{lang2024iseg} separates the rods, only one rod is aligned.

In the second case, the backrest of the hand-shaped chair statue is composed of four ``fingers''. A click on the pinkie is matched correctly to a vertex on the backrest of the 3D chair. However, while the entire backrest of the mesh is segmented, the 2D model marks only the pinkie region in the image. Thus, in both cases, the resulting image-shape segmentation matching is partial.

\section{Quantitative Evaluation} \label{sec:quantitative_evaluation}

To the best of our knowledge, a dataset with image-shape segment correspondence annotations is nowhere to be found. Thus, we constructed one ourselves from the PartNet dataset \cite{mo2019partnet}, as described in \cref{sec:fidelity}. PartNet is an annotated 3D part segmentation dataset, containing household objects, such as chairs, hats, etc. To create our cross-modality correspondence dataset, we sampled up to 20 shapes from the dataset's categories (fewer for categories with a smaller number of instances).

From each shape, we selected 10 vertices from a part at random, and rendered the shape from a set a views, with elevation of $\{\ang{-60}, \ang{30}, \ang{0}, \ang{30}, \ang{60}\}$, and azimuth of $\{\ang{0}, \ang{30}, ..., \ang{330}\}$, a total of $5 \cdot 12 = 60$ possible views. We randomly selected two of these views in which the vertex was visible. We discarded vertices that were not visible in all views and shape instances for which all 10 vertices were not visible.

\begin{figure*}[!t]
    \centering
    \newcommand{\plu}{0.5}
    \newcommand{\pld}{-1.7}
    \begin{overpic}[width=\linewidth, trim=0 0 10 50, clip]{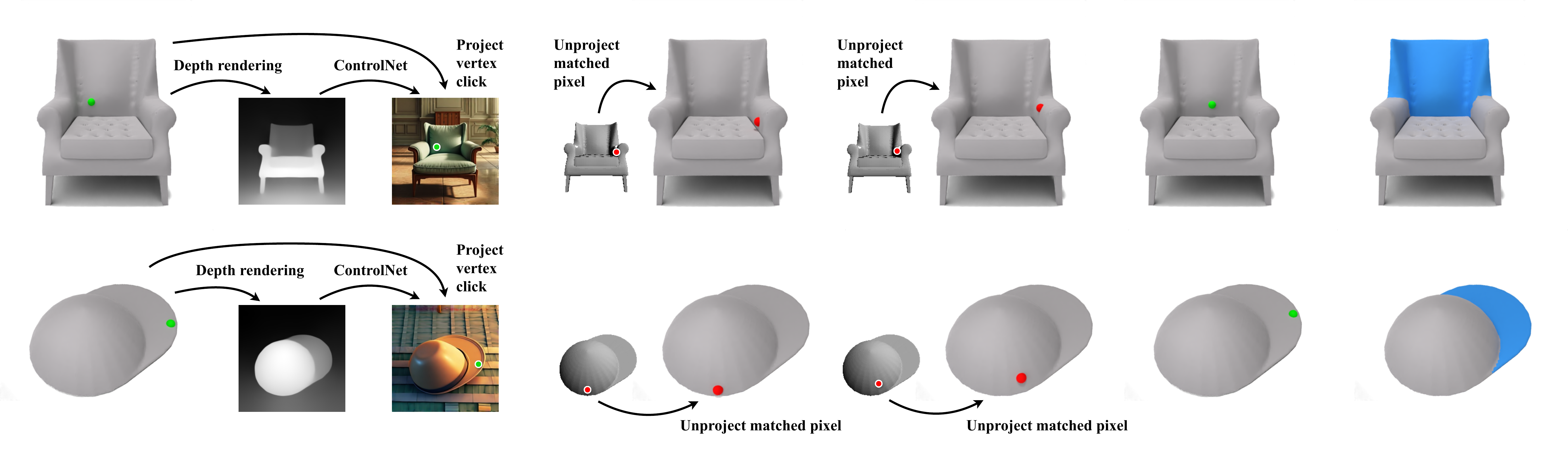}
    \put(3.8, \plu){\textcolor{black}{Clicked}}
    \put(4.5, \pld){\textcolor{black}{vertex}}
    \put(16.4, \plu){\textcolor{black}{Depth}}
    \put(14.9, \pld){\textcolor{black}{condition}}
    \put(24.5, \plu){\textcolor{black}{Generated}}
    \put(26, \pld){\textcolor{black}{image}}
    \put(40.5, \pld){\textcolor{black}{NBB \cite{aberman2018neural}}}
    \put(58, \pld){\textcolor{black}{DIFT \cite{tang2023imagediffusion}}}
    \put(74.5, \pld){\textcolor{black}{\ourmethod{} (ours)}}
    \put(89, \plu){\textcolor{black}{Ground-truth}}
    \put(92, \pld){\textcolor{black}{region}}
    \end{overpic}
    \vspace{-5pt}
    \caption{\textbf{Correspondence comparison on PartNet.} We show the generation process of the input image and 2D click (first three columns), the matched pixel by NBB and DIFT from the generated image to the rendered image of the shape and its unprojection to 3D (fourth to eighth columns), our matching vertex (ninth column) for the pixel click on the generated image, and the ground-truth shape region (tenth column) from which the vertex was selected (first column). Incorrect and correct correspondences are marked with red and green dots, respectively. While the baselines' correspondences are wrong, our method successfully finds a vertex within the ground-truth shape region.}
    \vspace{-2mm}
    \label{fig:partnet_comparison}
\end{figure*}
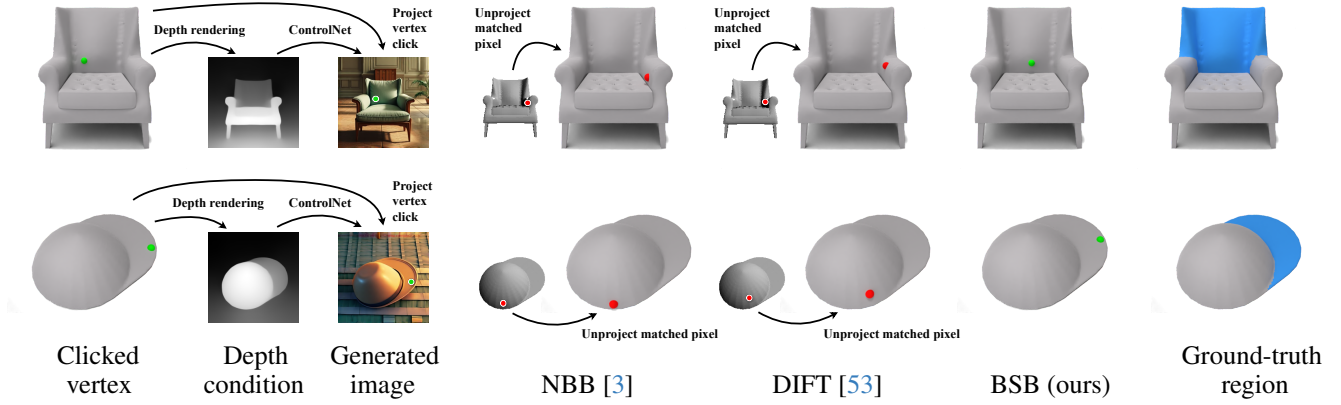

This process resulted in 265 shape instances, with a total of 2118 rendered images. For each of these views, we generated an image with ControlNet \cite{zhang2023adding} and projected the 3D click to the 2D view, as explained in \cref{sec:fidelity}, and visualized in \cref{fig:partnet_comparison}. To create naturally looking images, we used the prompt ``\textit{A picture of a \{category\} in realistic environments}''. Accordingly, our dataset contained 2118 image-shape pairs, with a ground-truth 3D region for the pixel click in the generated image.

This dataset was used for the quantitative evaluation of our method compared to the alternative zero-shot image correspondence baselines \cite{aberman2018neural, tang2023imagediffusion}, reported in \cref{tab:quantitative_evaluation}. \cref{fig:partnet_comparison} shows example results from the evaluation. As shown in the table and seen in the figure, the competitors fail to compute meaningful correspondences, whereas our method successfully matches the clicked pixel in the generated image to a vertex within the ground-truth region of the 3D shape, showcasing the superiority of our correspondence approach.

As another baseline, we examined the nearest vertex in the feature space as the match to the pixel click. This baseline achieved a success rate of 0.73. We note that since no existing dataset provides ground-truth annotations for image-shape correspondence, we performed the large-scale quantitative evaluation (\cref{tab:quantitative_evaluation}) on our synthetically generated cross-modality dataset, in which images are generated conditioned on the underlying shape geometry, as demonstrated in \cref{fig:partnet_comparison}.

In this simplified setting, \textit{where the image and shape share the same underlying structure}, nearest-neighbor matching performs well, as correspondence is easier to establish when the object type and structure align. In our target setting, where the image may exhibit substantial structural differences from the shape, \ourmethod{} significantly outperforms the NN baseline, as evidenced by the perceptual study detailed next in \cref{sec:perceptual_study}.

\section{Perceptual Study} \label{sec:perceptual_study}

\begin{figure*}[!t]
    \centering
    \newcommand{\pl}{-0.8}
    \begin{overpic}[width=\linewidth, trim=0 -30 260 195, clip]{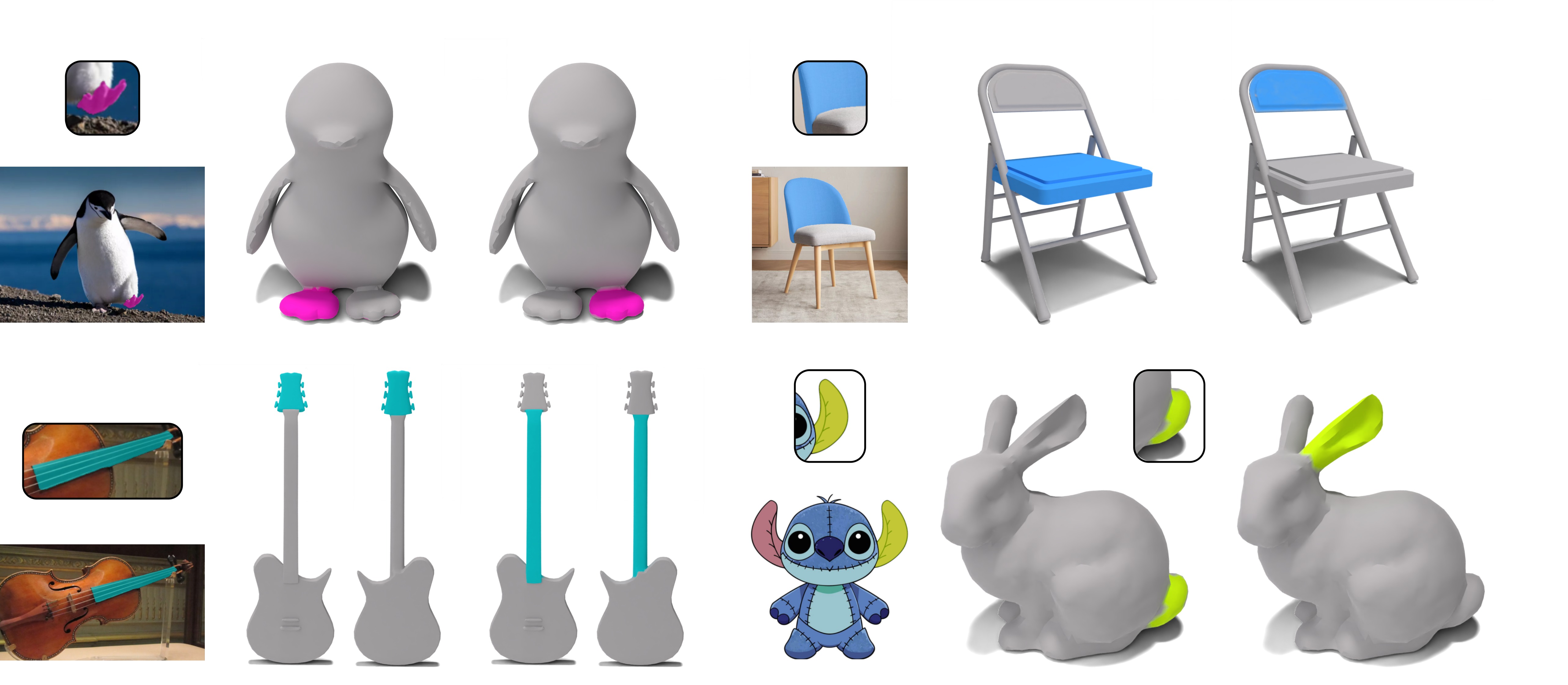}
    \put(16, \pl){\textcolor{black}{Nearest neighbor}}
    \put(34.8, \pl){\textcolor{black}{\ourmethod{} (ours)}}
    \put(65.5, \pl){\textcolor{black}{Nearest neighbor}}
    \put(88, \pl){\textcolor{black}{\ourmethod{} (ours)}}
    \end{overpic}
    \caption{\textbf{Nearest neighbor vertex selection.} Selecting the nearest neighbor vertex for a pixel click in the image leads to erroneous correspondences. In contrast, our \ourmethod{} overcomes the image-shape modality gap and finds correct matches.}
    \label{fig:nearest_neighbor}
\end{figure*}
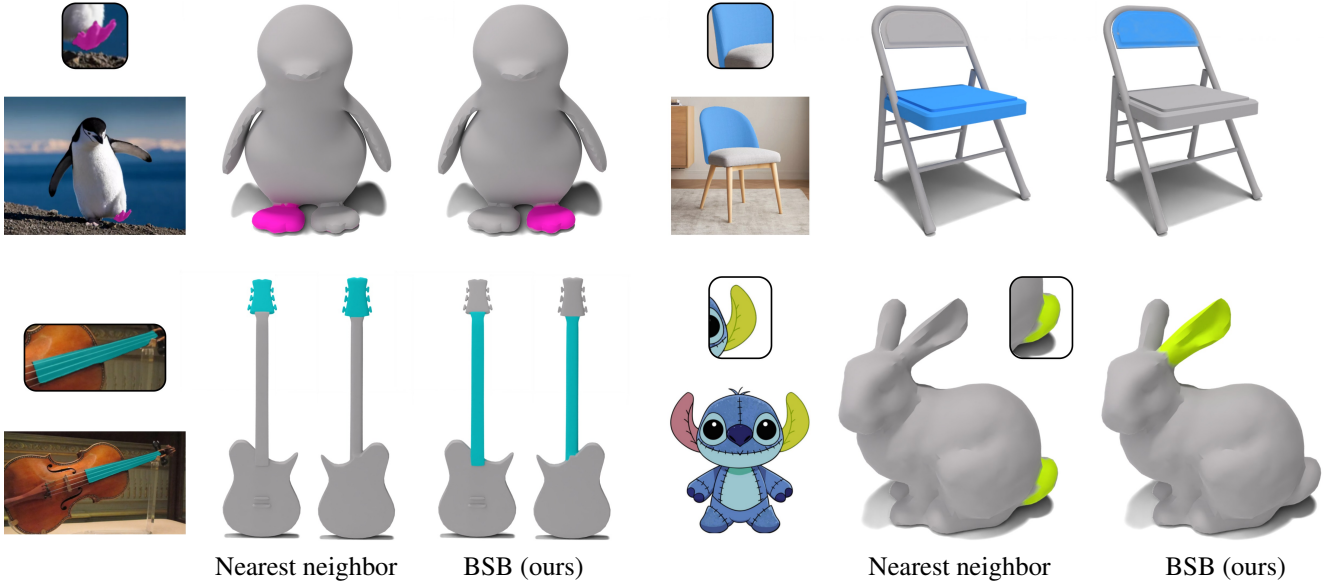

\ourmethod{} is not bound to a specific shape category or a given set of parts defined in a dataset. Thus, to evaluate the effectiveness of the flexible correspondences achieved by our method, where ground-truth labels do not exist, we conduct a perceptual user study. The survey included 31 participants who evaluated 20 image-shape pairs of various objects, including animals, humanoids, and household shapes, where the image and the shape differed in their structure or came from different domains. \cref{fig:comparison,fig:nearest_neighbor} present example image-shape pairs from the perceptual study.

For each pair, we showed the 2D and 3D corresponding segmentations for different methods and asked the participants to rate the effectiveness of the result on a scale of 1 to 5. The score 5 reflects a completely effective matching, where the correct part is segmented in 3D. When another part is segmented, the matching is partially effective, and when a wrong part is selected, or there is an empty segmentation, the correspondence is deemed completely ineffective, reflected by the score 1. We report the effectiveness scores averaged over the participants and the image-shape examples in \cref{tab:perceptual_study}. Our method is considered substantially more effective than the baselines, highlighting its high fidelity correspondences.

\cref{fig:nearest_neighbor} shows examples for selecting the nearest neighbor vertex for the pixel click in the image. Since there is a mismatch between the vision features of pixels and vertices, in this case, the pixel may be wrongly matched to a vertex in a different semantic region. For instance, the correspondence can be to another similar region in the shape, like the right penguin foot rather than its left one; to a vertex in an adjacent region, like the guitar’s head instead of its neck; or to a shape part that looks similar to the image segment but does not match it semantically, as the bunny's tail instead of its ear. In contrast, our \ourmethod{} considers different vertex candidates and selects the one that maps back to the segmented region in the image, enabling it to mitigate the modality shift and find correct correspondences.

\section{Ablation Test} \label{sec:ablation_test}

\begin{figure}[!t]
    \centering
    \vspace{-2.3mm}
    \includegraphics[width=\linewidth, trim=0 0 0 0, clip]{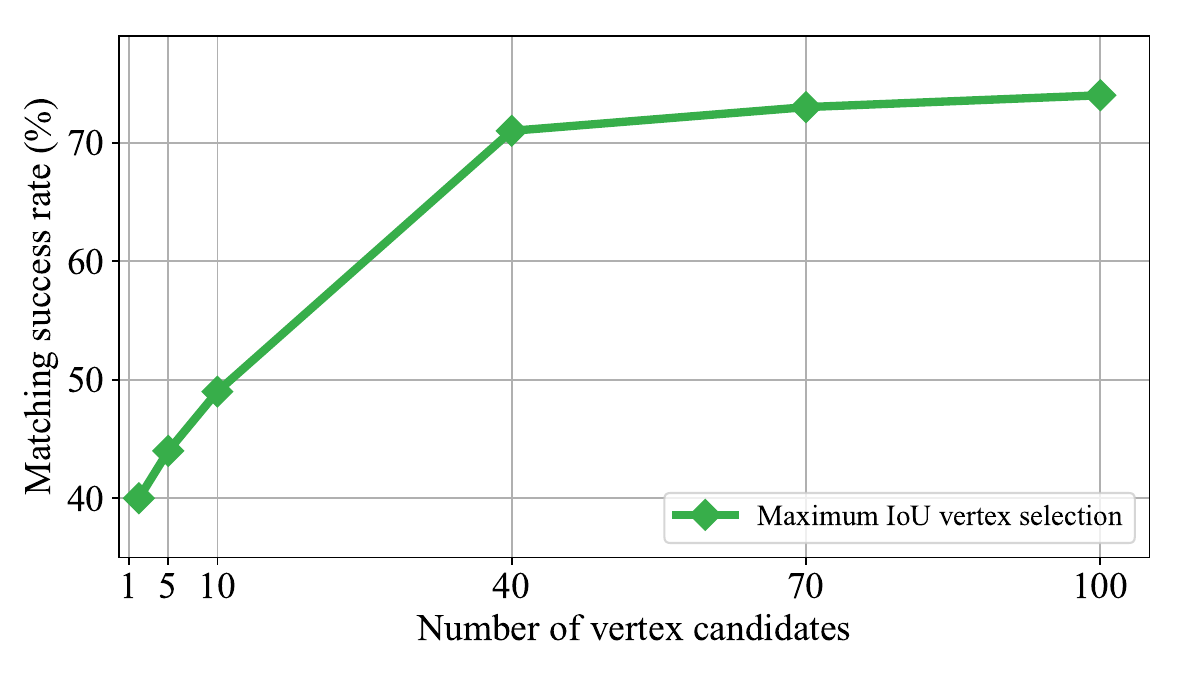}
    \vspace{-7mm}
    \caption{\textbf{Different number of vertex candidates.} We evaluate the matching success rate on PartNet for different values of vertex candidates. The performance starts to increase with a higher number of candidates and then saturates.}
    \label{fig:ablation_plot}
\end{figure}

We validate the design choices in our method with ablation experiments. First, we examine the influence of the number of vertex candidates $k$ on the correspondence success rate for the image-shape pairs from the quantitative evaluation discussed in \cref{sec:fidelity}. \cref{fig:ablation_plot} presents the results.

\begin{table}[!t]
\centering
\begin{tabular}{@{~} l @{~~~} c @{~~~} c @{~~~} c @{~~~} c @{~}}
\toprule
Method & NBB & DIFT & NN Baseline & \ourmethod{} (ours) \\
\midrule
Effectiveness $\uparrow$ & 2.75 & 2.74 & 3.26 & \textbf{4.63}  \\
\bottomrule
\end{tabular}
\vspace{-2pt}
\caption{\textbf{Perceptual user study.} We evaluate the image-shape correspondence effectiveness on a scale of 1 (completely ineffective) to 5 (completely effective). NN Baseline stands for selecting the nearest neighbor vertex in the feature space as the match for the pixel click. Our method is considered much more effective than the competitors.}
\label{tab:perceptual_study}
\end{table}

Due to the modality difference, there is a shift in the vision features of pixels and vertices. Thus, when $k$ is too small, a vertex from the corresponding ground-truth region is typically absent from the $k$ candidates of the clicked pixel, resulting in a low correspondence success rate. As $k$ is increased, additional candidates are considered, including ones from the ground-truth region, and a vertex from the correct 3D region is more frequently found by our \ourmethod{} mechanism, improving the success rate. Finally, the performance saturates at a success rate of 0.74 towards $k = 100$, leading to this configuration of $k$ in our method.

Additionally, we checked the case of randomly selecting a vertex for correspondence out of the $k=100$ candidates \vs choosing the one that maps to a pixel with the highest IoU with the image segmentation region, as proposed in our work. We have found that the former case results in a considerably lower matching success rate of 0.48, validating the effectiveness of the proposed selection procedure for the corresponding vertex.

\section{Implementation Details} \label{sec:implementation_details}

\noindent \textbf{Vision model distillation.} We train a multi-layer perceptron (MLP) to map each mesh vertex to a DINOv2-like feature vector of size $d_{vis} = 1024$. At the network's input, we use the vertex coordinates together with positional encoding of size 2048. The MLP contains 6 layers with 1024 neurons each. Each layer, except the last one, includes ReLU activation and layer normalization. For the last layer, we apply hyperbolic tangent activation without normalization.

For training the MLP, we render images of size $224 \times 244$ for 1000 random views of the mesh, where the angles were sampled randomly from the full elevation and azimuth range of $[-\ang{180}, \ang{180}]$ and $[\ang{0}, \ang{360})$, respectively. Each image was encoded by the DINOv2 ViT-L14 model to a tensor of size $w' \times h' \times d_{vis} = 16 \times 16 \times 1024$, and then interpolated to a spatial size of $64 \times 64$. Then, we rendered the predicted features for the mesh using a differentiable renderer into a tensor of size $64 \times 64 \times 1024$ and compared them to the reference DINOv2 features using a mean squared error loss. The network was trained with an ADAM optimizer for 3 epochs with a learning rate of 0.001. The entire distillation process, including rendering mesh views, encoding them, and training the network, takes only 3.5 minutes.

\smallskip
\noindent \textbf{Compared methods.} For the compared baselines \cite{aberman2018neural,tang2023imagediffusion}, we used the publicly available code bases released by the authors, with their recommended configuration.

\vspace{1mm}

\begin{algorithm}
\caption{Best Segmentation Buddies (\ourmethod{}) matching}
\label{alg:bsb_matching}
\begin{algorithmic}[1]

\Require Image $\Ical$ of size $(w,h)$, clicked pixel $p$, mesh $\Mcal$ with vertices $\Vcal$, 
2D segmentation model $\mathscr{F}^{2D}_{seg}$, 
2D vision model $\mathscr{F}^{2D}_{vis}$, 
distilled 3D MLP $\mathscr{F}^{3D}_{vis}$, 
number of vertex candidates $k$.

\Ensure Best segmentation buddy vertex $v_p$ (or \texttt{None}).

\State \textbf{Preprocess:} distill $\mathscr{F}^{2D}_{vis}$ into $\mathscr{F}^{3D}_{vis}$ from multi-view renders of mesh $\Mcal$.

\Statex \textbf{Feature extraction:}

\State $(M^{2D}_o, M^{2D}_p) \gets \mathscr{F}^{2D}_{seg}(\Ical, p)$
\State ${F'}_{vis}^{\Ical} \gets \mathscr{F}^{2D}_{vis}(\Ical)$
\State $F_{vis}^{\Ical} \gets \text{interpolate}({F'}_{vis}^{\Ical}, (w,h))$
\State $F_{vis}^{\Vcal} \gets \mathscr{F}^{3D}_{vis}(\Vcal)$

\Statex \textbf{Best segmentation buddy search:}

\State $s_{pv} \gets \cossim(F_{vis}^{\Ical}[p], F_{vis}^{\Vcal}[v])$ for all $v \in \Vcal$
\State $\mathcal{C} \gets$ top-$k$ vertices by $s_{pv}$

\State $maxIoU \gets 0$, $v_p \gets \texttt{None}$

\ForAll{$v' \in \mathcal{C}$}

    \State $q' \gets \argmax_{q \in M^{2D}_o}
    \cossim(F_{vis}^{\Vcal}[v'], F_{vis}^{\Ical}[q])$

    \If{$q' \notin M^{2D}_p$}
        \State \textbf{continue}
    \EndIf

    \State $M^{2D}_{q'} \gets \mathscr{F}^{2D}_{seg}(\Ical, q')$
    \State $iou \gets \mathrm{IoU}(M^{2D}_p, M^{2D}_{q'})$

    \If{$iou > maxIoU$}
        \State $maxIoU \gets iou$
        \State $v_p \gets v'$
    \EndIf

\EndFor

\If{$v_p = \texttt{None}$}
    \State \Return \texttt{None}
\Else
    \State \Return $v_p$
\EndIf

\end{algorithmic}
\end{algorithm}

\end{document}